\def\cvprPaperID{}
\def\confName{ICCV}
\def\confYear{2023}
\def\paperTitle{CHORUS: Learning Canonicalized 3D Human-Object Spatial Relations \\ from Unbounded Synthesized Images}

\def\authorBlock{
    Sookwan Han \qquad
    Hanbyul Joo \qquad \\
    Seoul National University \\
    {\tt\small \{jellyheadandrew, hbjoo\}@snu.ac.kr}
}

\newif\ifreview 
\newif\ifarxiv \newcommand{\arxiv}{\arxivtrue}
\newif\ifcamera 
\newif\ifrebuttal

\arxiv

\pdfoutput=1
\documentclass[10pt,twocolumn,letterpaper]{article}
\ifreview \usepackage[review]{cvpr} \fi
\ifarxiv \usepackage[pagenumbers]{cvpr} \fi
\ifrebuttal \usepackage[rebuttal]{cvpr} \fi
\ifcamera \usepackage{cvpr} \fi

\usepackage{graphicx}
\usepackage{amsmath}
\usepackage{amssymb}
\usepackage{booktabs}

\usepackage{times}
\usepackage{microtype}
\usepackage{epsfig}
\usepackage[table,xcdraw]{xcolor}
\usepackage[skip=5pt]{caption}
\usepackage{float}
\usepackage{placeins}
\usepackage{color, colortbl}
\usepackage{stfloats}
\usepackage{enumitem}
\usepackage{tabularx}
\usepackage{xstring}
\usepackage{cuted} 
\usepackage{multirow}
\usepackage{xspace}
\usepackage{url}
\usepackage{subcaption}
\usepackage{mathtools}
\usepackage[hang,flushmargin]{footmisc}
\usepackage[numbers,sort,compress]{natbib}
\usepackage[normalem]{ulem}  
\usepackage{graphicx}
\usepackage{amsmath}
\usepackage{amssymb}
\usepackage{comment}
\usepackage[title]{appendix}
\usepackage{cancel}
\usepackage{bbm}
\usepackage{subcaption}
\usepackage{afterpage}

\ifcamera \usepackage[accsupp]{axessibility} \fi 

\DeclareMathOperator*{\argmin}{arg\,min}

\newfloat{sepfigure}{p}{lop}
\floatname{sepfigure}{Figure}

\newcommand\blfootnote[1]{
  \begingroup
  \renewcommand\thefootnote{}\footnote{#1}
  \addtocounter{footnote}{-1}
  \endgroup
}

\ifarxiv  \fi

\newcommand*\xoverline[2][0.75]{%
    \sbox{\myboxA}{$\m@th#2$}%
    \setbox\myboxB\null
    \ht\myboxB=\ht\myboxA%
    \dp\myboxB=\dp\myboxA%
    \wd\myboxB=#1\wd\myboxA
    \sbox\myboxB{$\m@th\overline{\copy\myboxB}$}
    \setlength\mylenA{\the\wd\myboxA}
    \addtolength\mylenA{-\the\wd\myboxB}%
    \ifdim\wd\myboxB<\wd\myboxA%
       \rlap{\hskip 0.5\mylenA\usebox\myboxB}{\usebox\myboxA}%
    \else
        \hskip -0.5\mylenA\rlap{\usebox\myboxA}{\hskip 0.5\mylenA\usebox\myboxB}%
    \fi}
\makeatother

\newcommand{\R}[1]{{%
    \textbf{%
        \ifstrequal{#1}{1}{\textcolor{red}{R#1}}{%
        \ifstrequal{#1}{2}{\textcolor{blue}{R#1}}{%
        \ifstrequal{#1}{3}{\textcolor{magenta}{R#1}}{%
        \ifstrequal{#1}{4}{\textcolor{teal}{R#1}}{%
                           \textcolor{cyan}{R#1}%
        }}}}%
    }%
}}

\usepackage{xr-hyper}

\makeatletter
\newcommand*{\addFileDependency}[1]{
  \typeout{(#1)}
  \@addtofilelist{#1}
  \IfFileExists{#1}{}{\typeout{No file #1.}}
}

\makeatother

\usepackage[pagebackref,breaklinks,colorlinks]{hyperref}
\usepackage[capitalize]{cleveref}
\crefname{section}{Sec.}{Secs.}
\crefname{table}{Table}{Tables}
\crefname{figure}{Fig.}{Figs.}

\begin{document}

\title{\paperTitle}
\author{\authorBlock}
\maketitle

\begin{strip}
    \centering
    \vspace{-40pt}
    \includegraphics[width=1.0\linewidth, trim={0 0 0 0},clip]{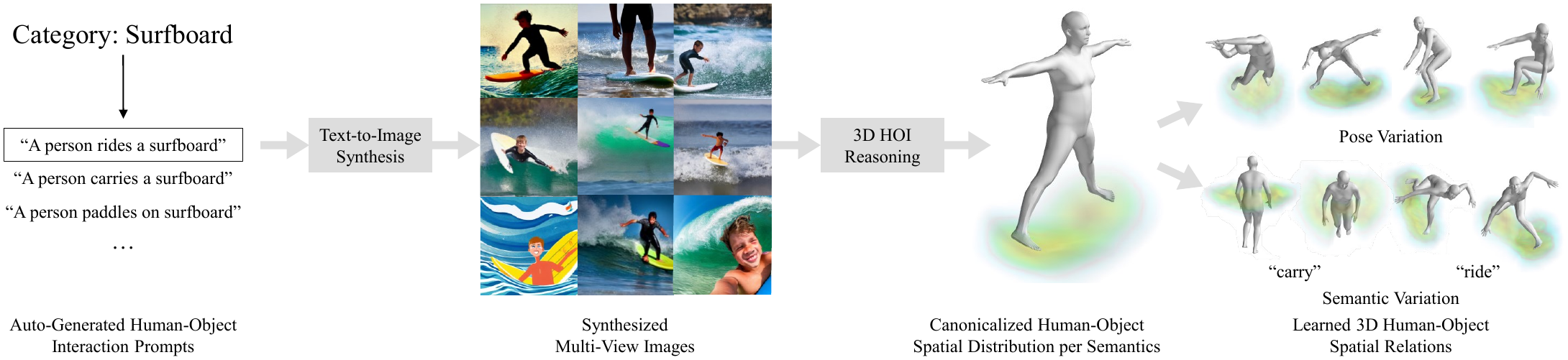}
    \captionof{figure}{\textbf{CHORUS}. Given an object category, our method, CHORUS, automatically learns human-object 3D spatial arrangements in a self-supervised way. Our method leverages a generative model to synthesize an unbounded number of images to reason about the 3D spatial relationship. As output, CHORUS produces the 3D spatial distribution in the canonical space, which can be deformed for any posed space.}
    \label{fig:teaser}
    \vspace{-5pt}
\end{strip}

\begin{abstract}
We present a method for teaching machines to understand and model the underlying spatial common sense of diverse human-object interactions in 3D in a self-supervised way. 
This is a challenging task, as there exist specific manifolds of the interactions that can be considered human-like and natural, but the human pose and the geometry of objects can vary even for similar interactions.
Such diversity makes the annotating task of 3D interactions difficult and hard to scale, which limits the potential to reason about that in a supervised way. One way of learning the 3D spatial relationship between humans and objects during interaction is by showing multiple 2D images captured from different viewpoints when humans interact with the same type of objects. 
The core idea of our method is to leverage a generative model that produces high-quality 2D images from an arbitrary text prompt input as an ``unbounded'' data generator with effective controllability and view diversity. Despite its imperfection of the image quality over real images, we demonstrate that the synthesized images are sufficient to learn the 3D human-object spatial relations.
We present multiple strategies to leverage the synthesized images, including (1) the first method to leverage a generative image model for 3D human-object spatial relation learning; (2) a framework to reason about the 3D spatial relations from inconsistent 2D cues in a self-supervised manner via 3D occupancy reasoning with pose canonicalization; (3) semantic clustering to disambiguate different types of interactions with the same object types; and (4) a novel metric to assess the quality of 3D spatial learning of interaction.
\blfootnote{Project Page: \href{https://jellyheadandrew.github.io/projects/chorus}{https://jellyheadandrew.github.io/projects/chorus}}
\end{abstract}

\vspace{-15pt}
\section{Introduction}
\label{sec:intro}
Humans interact with objects in specific ways. We wear shoes on our feet, a hat on our heads, and ride a bike by holding handles and putting two feet on the pedals.
While this common sense regarding the 3D arrangements of the way we interact with objects is known to us, teaching such things to robots and machines is challenging, requiring numerous variations in diverse human-object interactions in 3D.

As providing 3D supervision by manually annotating various cases is hard to scale, an alternative way of teaching such things is by showing 2D photos from the Internet containing the interaction with the same object from many viewpoints. A text-based image retrieval (e.g., Google Image Search) can be an option to crawl many images from a text description similar to NEIL~\cite{neil}. However, this approach fundamentally suffers from several obstacles such as (other than the challenges in 3D spatial relation reasoning): 
(1) viewpoint variations are insufficient and hard to control; (2) the number of related images decreases drastically as the compositionality of the prompt increases; and (3) the images are often biased due to commercial websites. 

\begin{figure}[t]
\centering
\includegraphics[width=1.0\columnwidth]{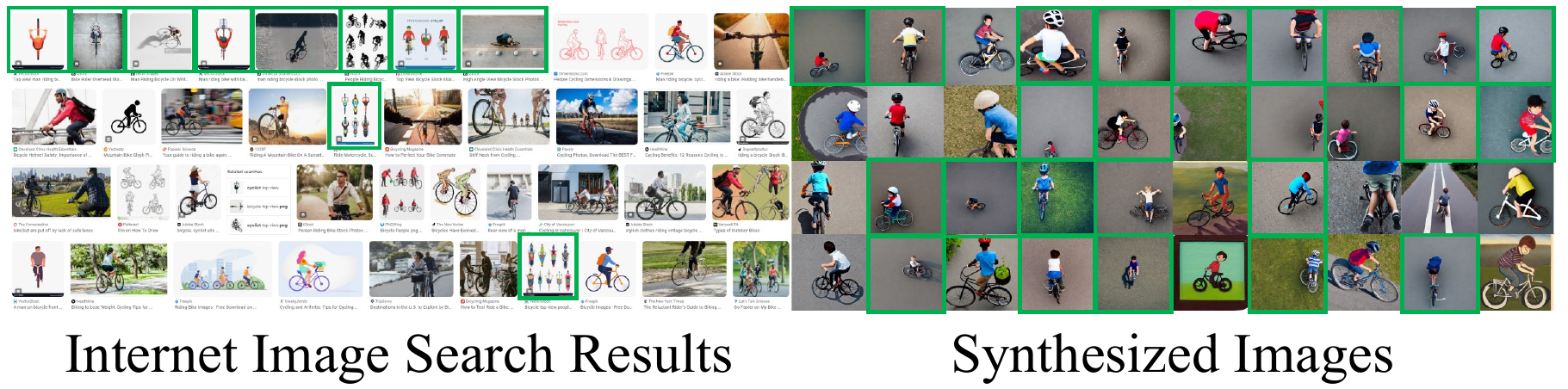}
\caption{\textbf{Internet-crawled images vs. synthesized images.} Under the prompt ``A person is riding a bicycle, top view'', synthesized images exhibit superior fidelity to the intended ``top view'' perspective compared to internet-crawled images. The correct viewpoint image is indicated by the green bounding box.}
\label{fig:comparison_web}
\vspace{-15pt}
\end{figure}

In this paper, we present a novel idea of leveraging a text-conditional generative model~\cite{diffusion_ldm} as a controllable data producer in synthesizing ``unbounded'', ``multi-view'', ``diverse'' images to learn the 3D human-object spatial relations in a self-supervised manner.
Despite its imperfectness in quality, we observe that the synthesized images from generative models are more suitable for our objective, as the generative model effectively links desired semantics of human-object interaction (HOI) described in natural language. Our synthesis-based approach allows better controllability in obtaining images for spatial relation learning, providing more relevant images from diverse viewpoints. See the examples in Fig.~\ref{fig:comparison_web}. 

Nonetheless, inferring 3D spatial knowledge in human-object interaction from in-the-wild 2D image collections is still a non-trivial problem due to the inconsistency and ``wildness'' among synthesized images. In particular, a method should handle the following challenges: (1) semantic variation: given a category, there can be various semantic situations of human-object interaction; thus, the spatial distribution of object may vary significantly as semantics vary; (2) human pose variation: even assuming the same semantic situation, human postures can vary as diverse actions and pose are available in the same situation;  (3) intra-class variation: object can exist in various forms (even if same category); and (4) visual variance: when learning from 2D cues, the visual properties (e.g., illumination, camera) may vary for even the same 3D arrangement, making it difficult to localize and extract 2D cues.

To this end, we present a self-supervised method to learn the spatial common sense of diverse human-object interactions in 3D for arbitrary object categories without any human annotations. We present several novel components to achieve the goal, including (1) automatic prompt generation for diverse view and semantic variations via chatGPT~\cite{chatgpt}, (2) outlier filtering strategies, (3) automatic camera view calibrations by using a human as an anchor, (4) accumulating spatial interaction cues from inconsistent multi-view 2D knowledge with varying human pose, object geometry, and (5) clustering for various semantic human-object interaction types.
The output of our method can be considered as \textit{possible occupancy distributions} of the category-specified 3D object relative to the human body in a canonical person-centric space regarding the intra-class object variation. We demonstrate the efficacy of our method on various object categories and human-object interaction types, as shown in Fig.~\ref{fig:teaser} and Fig.~\ref{fig:qual_eval_various_categories}.
As the first in this direction, we also introduce a new metric, namely \textit{Projective Average Precision (PAP)}, to quantify the quality of 3D spatial inference outputs. 
Our contributions are summarized as follows: (1) the first method to leverage a generative text/image model for 3D human-object spatial relation learning, including automatic prompt generation, outlier filtering, and 3D viewpoint estimation via estimated 3D humans; (2) a framework to reason about the 3D spatial relations from inconsistent 2D cues in a self-supervised manner via 3D occupancy reasoning with pose canonicalization; (3) semantic clustering to disambiguate different types of interactions with the same object types; (4) a novel metric to assess the quality of 3D spatial learning of interaction.
\section{Related Work}
\label{sec:related}
\noindent \textbf{Text-to-Image Synthesis with Diffusion Models.}
Recent approaches in text-to-image synthesis show great performance by leveraging emerging diffusion models~\cite{diffusion_sohldickstein_2015,diffusion_ncsn, diffusion_iddpm, diffusion_edm, diffusion_ddim, diffusion_ldm, diffusion_vesde_vpsde}. Diffusion models are a class of generative models that ``noises'' the data from training distribution and learn to ``denoise'' the perturbed images at arbitrary noise scale.
Diffusion models are likelihood-based generative models and are known to show high mode coverage and generate high-fidelity images. One drawback is the slow inference time due to multiple iterations of the denoising process, which can be mitigated by recent approaches~\cite{diffusion_ddim, diffusion_edm, diffusion_pseudonumerical, diffusion_ldm}.
Diffusion models enable text-conditional image generation via utilizing CLIP~\cite{clip} embeddings~\cite{diffusion_glide, diffusion_dalle2, diffusion_ldm} or large-language-model encodings \cite{diffusion_imagen}. In practice, classifier guidance~\cite{diffusion_diffusionbeatgans} or classifier-free guidance~\cite{diffusion_cfg} is applied at inference steps further to enhance the quality and text-coherency in trade-off with diversity.

\noindent \textbf{Generative Models as Data Producer.}
Similar to our approaches, there have been extensive approaches to leverage generative models as data generators.
Prior approaches utilize GANs~\cite{generative_model_as_data_producers_gan, generative_model_as_data_producers_gan2, generative_model_as_data_producers_gan3} to synthesize data for tasks including classification~\cite{generative_model_as_data_producers_classification, generative_model_as_data_producers_classification1, generative_model_as_data_producers_classification2, generative_model_as_data_producers_classification3, generative_model_as_data_producers_classification4}, 3D vision~\cite{generative_model_as_data_producers_3dvision, generative_model_as_data_producers_3dvision2, generative_model_as_data_producers_3dvision3, generative_model_as_data_producers_3dvision4}, 2D segmentation~\cite{generative_model_as_data_producers_segmentation1, generative_model_as_data_producers_segmentation2, generative_model_as_data_producers_segmentation3}, dense visual alignment~\cite{generative_model_as_data_producers_gangealing}; or diffusion models~\cite{diffusion_ddpm, diffusion_ldm} for data augmentation~\cite{generative_model_as_data_producers_3dvision4_usediffusion_augmentation} or as synthetic data for few-shot learning~\cite{generative_model_as_data_producers_3dvision4_usediffusion_synthdataset}. One major challenge of such methods is that generative models output ``free'' and ``uncontrolled'' images. Ali \textit{et al.}~\cite{generative_model_as_data_producers_genmodelsdatasourcemultiviewrepr} presents a method to generate multi-view images for representation learning by applying transformations on the latent vectors of the generative models, similar to our approach.

\noindent \textbf{Learning 3D Spatial Arrangement between Human and Object.} 
Prior work models 3D human-object interaction for each specified interaction type and input. Approaches include inferring hand-object interaction from 3D~\cite{hoi_contactgrasp,hoi_grab,hoi_toch}, 2.5D such as heatmap~\cite{hoi_contactdb,hoi_contactpose}, or provided images~\cite{hoi_ganhand,hoi_useforceluke,hoi_learn_joint_hands_manipulated_objects,hoi_grasping_field,hoi_cpf}. These methods only model hand-object relationships and cannot predict the full body. The methods that model full-body interactions include generating/reconstructing 3D humans from 3D scene constraints~\cite{hoi_posa,hoi_pigraphs,hoi_place,hoi_couch}, capturing interaction from multi-view camera \cite{hoi_behave,hoi_neuralhofusion,hoi_neuralfreeviewpointhoi}, or reconstructing 3D scene from human-scene interaction type~\cite{hoi_hoaware_placement}, or human-human interactions~\cite{hoi_3dreconhuman}. 
Also, there exist approaches to model contacts~\cite{hoi_holisticscenepp,hoi_capture_infer_dense_body,hoi_longterm_humanmotion,hoi_motion_from_mono_video,hoi_imapper,hoi_contact_from_mono_video}. PHOSA~\cite{hoi_phosa} reconstructs 3D human and object from a single image; however, it requires manual labeling for the human-object interaction region, which limits the generalizability of the approach. CHORE~\cite{hoi_chore} additionally models contact via distance between human and object by leveraging implicit surface learning to fit parametric SMPL~\cite{smpl} human and template object mesh. 
Unlike previous methods, our method does not require any object templates or supervision.
\section{Method}
\label{sec:method}

\begin{figure*}[t]\centering
\includegraphics[width=\linewidth, trim={0 0 0 0},clip]{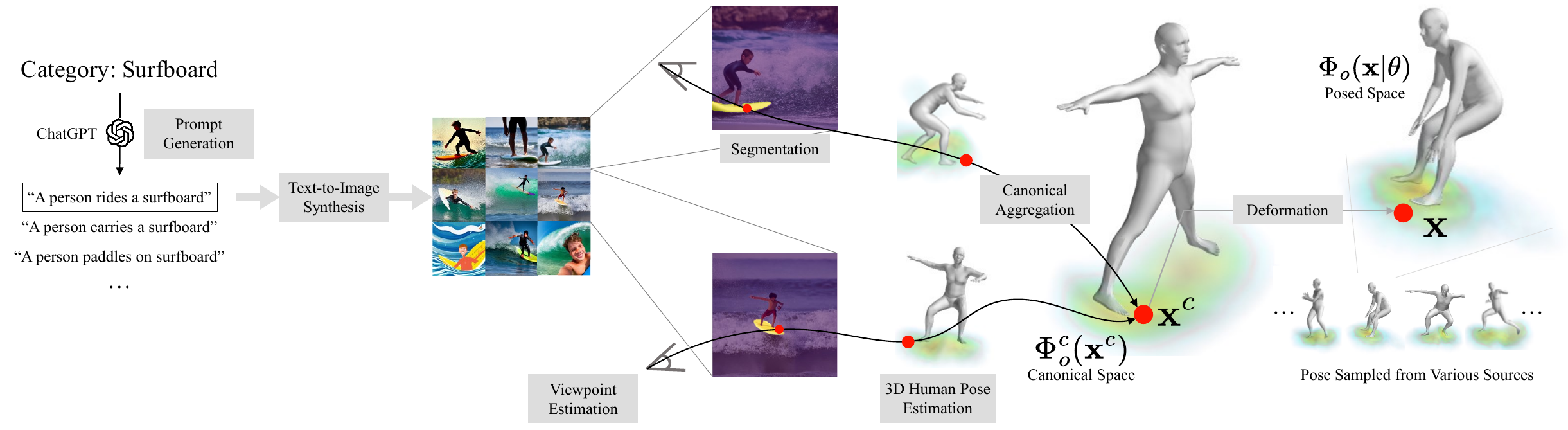}
\captionof{figure}{\textbf{Method overview.} Our method starts with generating prompts for human-object relationships within a specific object category. These prompts produce a multitude of images, incorporating HOI semantics. After applying filtering strategies to eliminate outliers, the chosen images are aggregated in canonical 3D space using a canonicalization approach. The resulting distribution can be flexibly adapted to different human postures.}
\label{fig:overview}
\vspace{-10pt}
\end{figure*}

\subsection{Overview}
\label{subsec:overview}
Our method extracts 3D spatial knowledge of human-object interactions (HOI) by modeling object location as an occupancy probability distribution $\Phi_o$ when the category of the object $o$ is given.
The spatial relationship during HOI can vary even for the same object category due to the variation of human state (e.g., pose, body height and body shapes) and types of interactions (e.g., we can ride or hold a surfboard). Thus, we model the 3D spatial probability distribution in pose-conditioned and type-conditioned manners:
\begin{equation}
     \Phi_o( \mathbf{x} | \theta, s) \in [0,1],
    \label{eq:HOI_distribtuion}
\end{equation}
where $\mathbf{x} \in \mathbb{R}^3$ is a 3D location of object occurrence in the ``pose-deformed'' space, $\theta$ is the 3D state of a person in terms of pose and shape variation, and $s$ is a type of human-object interaction among possible discrete variations $\mathbf{S}$.
We represent $\Phi_o$ in an explicit voxel space with $48^3$ resolutions, where the probability shows the likelihood that the specific location is occupied by the object $o$ during interactions. The human state $\theta \in \mathbb{R}^{24 \times 3}$ is parameterized by SMPL~\cite{smpl} pose parameters for 24 joints to represent the condition of 3D human pose variations.
$\mathbf{S}$ represents plausible types of interactions for an object $o$, which is used for clustering distributions as described in Sec.~\ref{subsect:clustering}. Specifically, $\mathbf{S}$ considers the semantics described by (1) natural language prompts and (2) contacted parts between humans and objects. For example, holding a surfboard and riding a surfboard should be considered as different interactions, despite the same object category, which is the objective to have $\mathbf{S}$ as the input for our $\Phi_o$.

As a way to learn the 3D spatial distribution from inconsistent 2D image data, our method aggregates the HOI cues in a canonical space:
\begin{equation}
     \Phi^c_o( \mathbf{x}^c | s) \in [0,1],
\label{eq:HOI_distribtuion_canonical}
\end{equation}
where the $\mathbf{x}^c$ is a 3D location at the canonical space. 
The canonical space is defined by the rest-pose of SMPL, as shown in Fig.~\ref{fig:overview}, where we put zero rotations for most joints except hips; $ \pm \pi/6$ z-axis rotation on the left and right hips, respectively; following the approach proposed by SNARF~\cite{snarf} (we empirically find that it is advantageous to keep a sufficient distance between legs).
Note that this canonical distribution $\Phi^c_o$ is independent of the 3D human pose state $\theta$, in contrast to Eq.~\ref{eq:HOI_distribtuion}.
Our 3D spatial reasoning in the canonical space is inspired by the recently emerging approaches to building animatable 3D avatars~\cite{snarf, gdna, scanimate, selfrecon}, where cues from multiple 3D scans or views from different postures are aggregated in the canonical space. Yet, our framework is much simpler without requiring precise linear-blend skinning estimation since our target is reasoning the approximate object locations w.r.t. human rather than high-fidelity 3D human surface estimation. 
By reasoning in the canonical space, we can handle the inconsistency and variation of synthesized multiview images in HOIs.
To this end, our framework provides a warping function to convert back and forth between $\Phi^c_o$ and $\Phi_o$, and our learned 3D spatial distribution can be applied to any 3D human postures to guess potential object locations as shown in Fig.~\ref{fig:qual_eval_various_categories}.
As an interesting key idea for generating multi-view image collections for 3D spatial relation learning, we use a SOTA diffusion model~\cite{diffusion_ldm} that can synthesize realistic images from text prompt inputs.
We refer readers to Fig.~\ref{fig:overview} for an overview of our method, and Supp. Mat.~\ref{appendix:promptgeneration}$\sim$\ref{appendix:selectiveaggregationviasemanticclustering} for more details on each component.

\subsection{Dataset Generation}
\label{subsec:synthesize_images}
\noindent \textbf{Prompt Generation.} We aim to produce various text prompts that describe the diverse semantics of human-object interaction on a target object $o$, which a text-conditioned diffusion model~\cite{diffusion_ldm} can then take for image generation. In particular, we want to obtain visual cues on various human-object interactions with the target object $o$ with many viewpoints for 3D reasoning.

\begin{figure}[t]
\centering
\includegraphics[width=1.0\columnwidth]{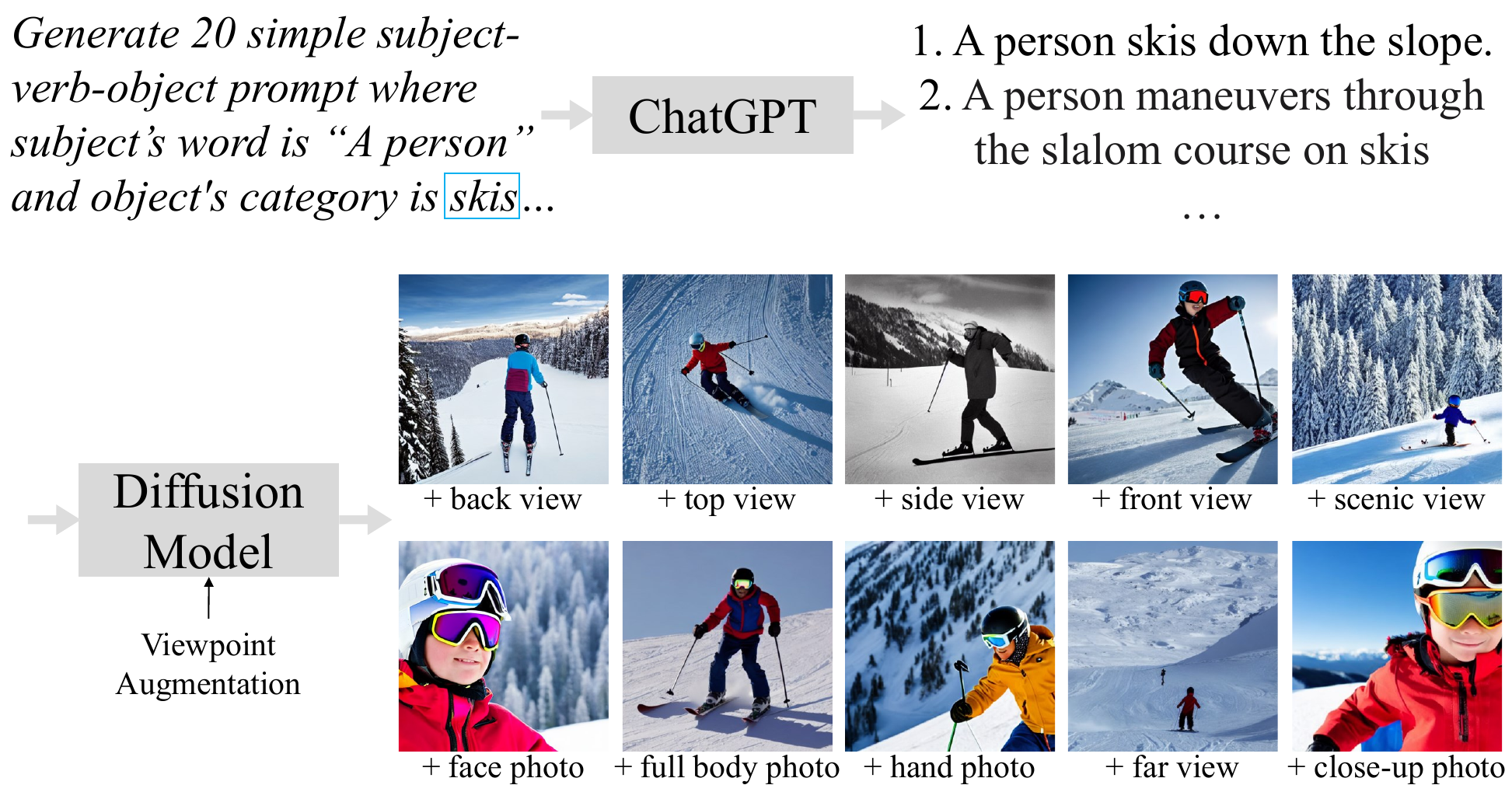}
\caption{\textbf{Viewpoint augmentation.} Examples of prompts produced by ChatGPT and images synthesized via diffusion with viewpoint augmentation.}
\label{fig:viewpoint_augmentation}
\vspace{-10pt}
\end{figure}

Towards an automatic process, we present a solution by leveraging the ChatGPT~\cite{chatgpt}, an instruction-tuned large language model, to produce a set of plausible prompts automatically. Specifically, we give the input query sentence as shown in Fig.~\ref{fig:viewpoint_augmentation}, where we can simply replace the object category (e.g., skis as shown in a blue box) for any target objects. As shown, ChatGPT produces various related output prompts. 
We empirically observe that our solution with ChatGPT is much more efficient with higher quality, compared to the possible alternative way of directly generating prompt sentences via exhaustive combinations of a set of subjects and verbs, which may produce awkward expressions (e.g., ``wear a bike'').
Furthermore, to encourage similar scenes from diverse viewpoints as much as possible, we also present a strategy to control viewpoints of the synthesized images by augmenting view conditions on each prompt produced from ChatGPT, such as ``back view'', ``side view'', and ``far view''. As shown in Fig.~\ref{fig:viewpoint_augmentation}, we demonstrate this viewpoint augmentation on prompts is advantageous in producing scenes from diverse views, while it is only partially guaranteed that the outputs follow the instructed viewpoints due to the imperfection of generative models. We generate 3 to 20 prompts per category, where each is augmented with 22 different auxiliary prompts to control viewpoints.

\noindent \textbf{Synthesizing Text-Conditioned Images via Diffusion.} Given a set of produced prompts regarding a human-object interaction, we link these natural language descriptions to visual entities by synthesizing images via an off-the-shelf latent diffusion model~\cite{diffusion_ldm}, which has great power in high-fidelity image generation with high mode coverage. Examples are shown in Fig.~\ref{fig:viewpoint_augmentation}.
Our strategy enables us to produce an unbounded number of diverse scenes containing the desired human object interaction with the target object category. Notably, the same prompt can be used multiple times by simply changing the initial random latent, resulting in different outputs.
We create $5000\sim 90000$ images per object category, evenly distributed for each given prompt within the category.

\noindent \textbf{Filtering.} Despite the efficacy of using the diffusion model for an image producer, the collected images are still in a wild status, making it challenging to use them directly for 3D spatial HOI learning. Thus, we first apply a series of filtering strategies to retain useful images only.
Notably, the unbounded nature of our data generation allows us to use tough thresholds to filter out useless image samples aggressively. 
We use the following criteria in determining ``valid'' image samples: (1) the image contains a single human for efficient localization without ambiguity; (2) the image contains the target object; (3) the human and the target object should be close enough with sufficient overlapping in the bounding boxes (with a certain threshold); and (4) the whole torso part of the human should be visible.
We use an off-the-shelf object detector~\cite{pointrend, detectron2} for the bounding box detection and instance segmentation. 
To check the torso's existence, we use a 2D keypoint detector~\cite{hrnet, darkpose, mmpose2020} and check whether shoulders and hips are within the image boundary and their detection confidences are above certain thresholds.

\subsection{Pose Canonicalized Aggregation}
\label{subsec:pose_canonicalization}
A large collection of images obtained from our approach can provide cues for 3D spatial HOI reasoning. As its first step, we estimate camera viewpoint by using the estimated 3D human orientation as an anchor. Then, we aggregate 2D object mask cues from each view to estimate 3D occupancy distribution, where we apply pose canonicalization to handle pose variations among synthesized images.

\begin{figure}[t]
\centering
 \begin{subfigure}[t]{0.57\columnwidth}
         \centering
    \includegraphics[width=1.0\columnwidth]{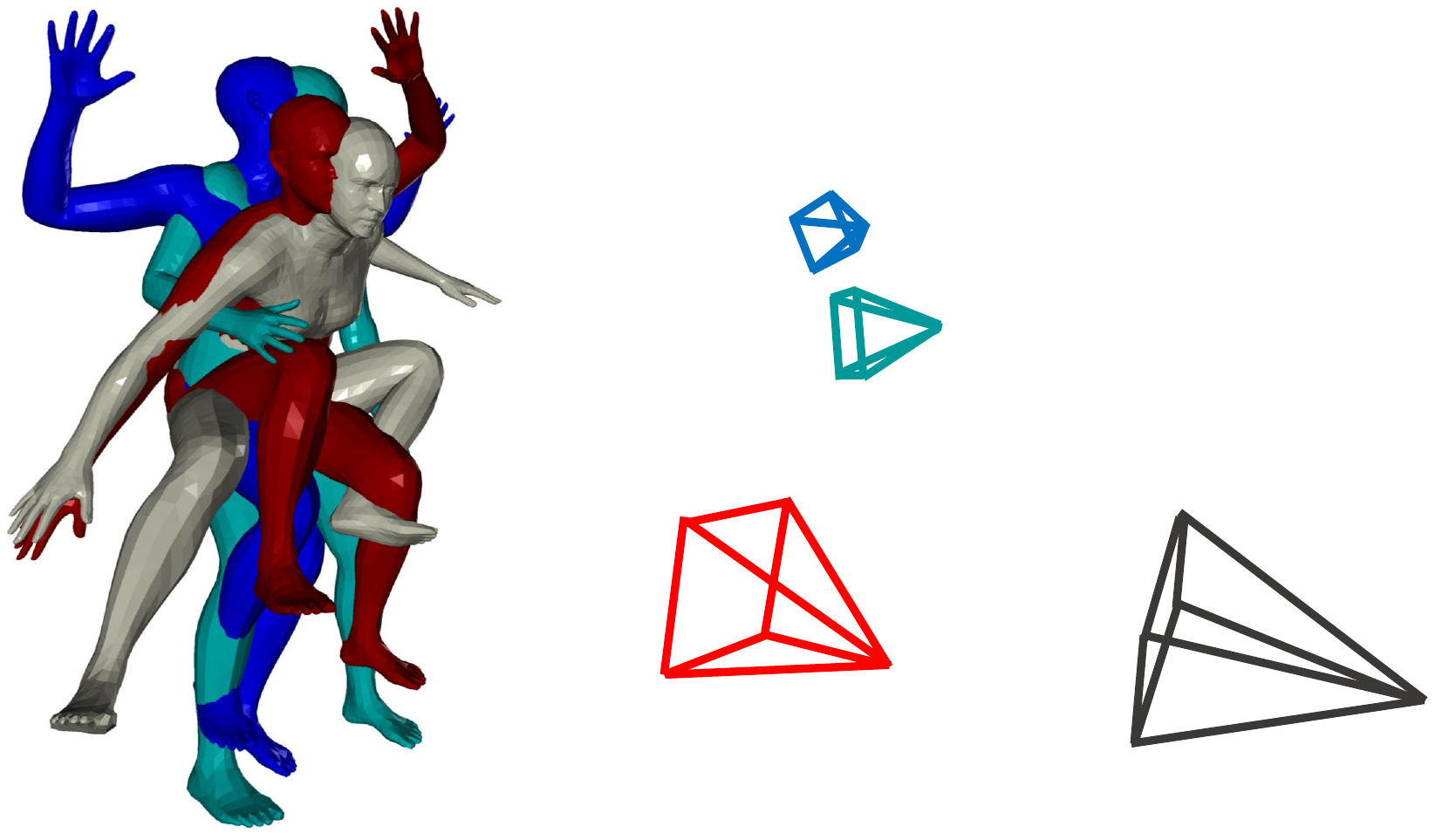}
    \caption{Calibrated cameras and humans.}
         \label{subfig:calibrated_3d}
\end{subfigure}
\begin{subfigure}[t]{0.38\columnwidth}
         \centering
    \includegraphics[width=1.0\columnwidth]{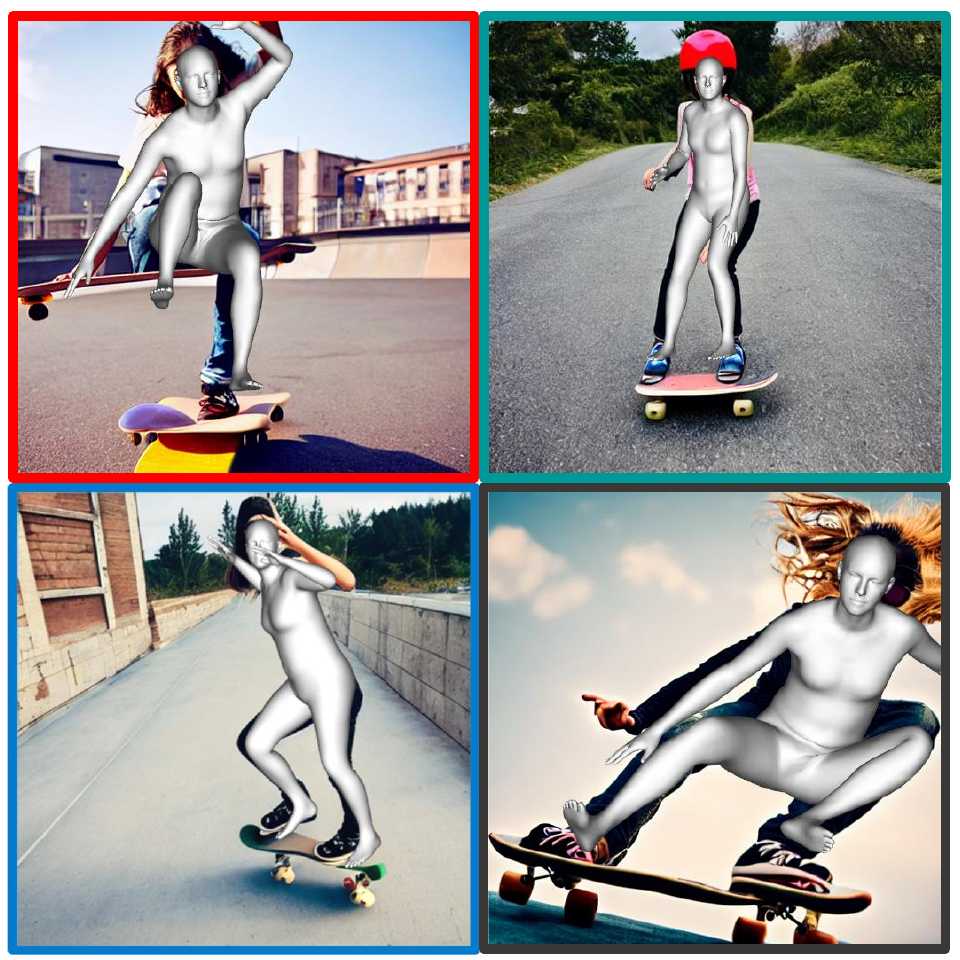}
    \caption{Synthesized images.}
         \label{subfig:calibrated_2d}
\end{subfigure}    
\caption{\textbf{Human-anchored camera calibration.} Predicted humans and calibrated cameras for synthesized images.
}
\label{fig:pose_diversity_for_calibrated}
\vspace{-10pt}
\end{figure}

\noindent \textbf{Viewpoint Estimation via 3D Human Pose Estimation.} The orientation of the human body in the scene can provide a clue to estimate the relative camera viewpoint w.r.t the human. For this purpose, we use an off-the-shelf monocular 3D human pose estimator~\cite{frankmocap}, which outputs the 3D global orientation of humans along with 3D joint orientations in SMPL~\cite{smpl} parameterzation. Specifically, given an image $\mathbf{I}$ and a human bounding box $\mathbf{B} \in \mathbb{R}^4$, a 3D pose regressor $\mathcal{F}_{\text{pose3d}}$ outputs:
\begin{equation}
    \label{eq:pose3d}
    \{\mathbf{\phi}, \mathbf{\theta}, \mathbf{\beta}, \pi, \mathbf{j}\} = \mathcal{F}_{\text{pose3d}}(\mathbf{I}, \mathbf{B}),
\end{equation}
where $\mathbf{\phi} \in \mathbb{R}^{3}$ is the global orientation of the human defined in the camera-centric coordinate, $\mathbf{\theta} \in \mathbb{R}^{23\times3}$ is 3D joint angles (excluding the pelvis root) in angle axis,  $\beta\in\mathbb{R}^{10}$ is shape parameters, and $\pi \in \mathbb{R}^3$ is the weak-perspective camera parameters, and $\mathbf{j}\in\mathbb{R}^{24\times2}$ is the projected 2D joint locations on image space. We then convert $\pi$ and $\phi$ into a perspective camera model $\Pi$  in the person-centric coordinate system by minimizing the distance between projected SMPL joints via $\pi$ and $\Pi$. To this point, all images are ``calibrated'' in a common 3D space, where the SMPL humans are aligned and centered at the origin (i.e., pelvis is set as origin), as depicted in Fig.~\ref{fig:pose_diversity_for_calibrated}.

\noindent \textbf{3D Occupancy Estimation via Human Pose Canonicalization.} We represent spatial 3D HOI reasoning via the occupancy probability field $\Phi^c_o$ as defined in Eq.~\ref{eq:HOI_distribtuion_canonical}.
For brevity, we may first consider a holistic distribution $\Phi^c_o( \mathbf{x}^c) \in [0,1]$, which does not consider semantics and returns a marginalized distribution for the most probable HOI.
Given the virtually calibrated multi-view images we processed above, we compute $\Phi^c_o$ via visual hull reconstruction using the 2D object segmentation masks obtained in the filtering stage.
However, as an issue, the 3D human poses may not be consistent across views, as shown in Fig.~\ref{fig:pose_diversity_for_calibrated}, making it challenging to apply visual hull reconstruction directly.

To handle this issue, we consider a canonical space defined via the rest pose of the SMPL model and compute the conversion between this canonical space and the pose-deformed space corresponding to each view. This process is inspired by the deformable object reconstruction~\cite{dynamicfusion} or the animatable human reconstruction pipelines~\cite{snarf, gdna, scanimate, selfrecon}, where SMPL body posture provides the guidance to convert between two spaces.
Unlike the 3D human modeling approaches that use MLPs for neural skinning~\cite{snarf, gdna, scanimate}, our goal is to enable the 3D voting to the corresponding 3D canonical volume space by warping into pose-deformed volume space and checking the means of 2D mask occupancies. Thus, we take a simple strategy to find the mapping between the canonical and posed spaces.
Specifically, we define Linear Blending Skinning (LBS) weights for 3D point $\mathbf{x}^c$ in the canonical space (we use $48^3$ voxel grid) by averaging the ones from $\mathbf{k}$-nearest neighbor SMPL vertices with inverse distance weights, similar to SelfRecon~\cite{selfrecon}. That is:
\begin{equation}
    \omega(\mathbf{x}^c) = {{\sum\limits_{i\in \mathbf{N}_{\mathbf{k}}(\mathbf{x}^c)} {w_i ~/~{\|\mathbf{x}^c-\mathbf{v}_i\|}}}\over{\sum\limits_{i\in \mathbf{N}_{\mathbf{k}}(\mathbf{x}^c)} {1 ~/~ {\|\mathbf{x}^c-\mathbf{v}_i\|}}}}
    \label{eq:lbs_nearest_neighbor}
\end{equation}
where $\mathbf{N}_{\mathbf{k}}(\mathbf{x}^c)$ is a set of $\mathbf{k}$-nearest neighbor vertex indices on SMPL mesh, $w_i\in \mathbb{R}^{24}$ and $\mathbf{v}_i\in\mathbb{R}^3$ are each associated LBS skinning weights and location of $i$-th vertex in SMPL. 
Intuitively, the 3D space is transformed by following the motion of the closest SMPL vertices.
Additionally, assuming the monotonic decrease of the effect of bone transformation as $\mathbf{x}^c$ is far from the SMPL mesh model, we encourage ``zero deformations'' for the far-away points by decreasing the effect of LBS with skinning weight adjustments by simply applying weighted sum between $\omega(\mathbf{x}^c)$ and LBS skinning weights for pelvis $\mathbf{e}_1\in\mathbb{R}^{24}$ which has no effect on deformation.
Refer to Supp. Mat.~\ref{appendix:3doccupancyestimationviahumanposecanonicalization} for more details.

Given computed LBS weights, the warping from the canonical space $\mathbf{x}^c$ to a pose-deformed space $\mathbf{x}$ with SMPL pose $\mathbf{\theta}$ can be performed by following the forward-kinematics pipeline of SMPL skeletal hierarchy:
\begin{equation}
    \label{eq:forwardkinematicpipeline}
    \mathbf{x} = \mathcal{W}(\mathbf{x}^c) =  
    \sum_{j=1}^{n_b} \omega_j(\mathbf{x}^c) \cdot \mathbf{B}_j(\theta_j) \cdot \mathbf{x}^c,
\end{equation}
where $\omega_j$ denotes the $j$-th LBS weights, $\mathbf{B}_j(\theta_j)\in\text{SE}(3)$ represent $j$-th bone's global 3D transformations.

Finally, the 3D occupancy aggregation can be performed by warping the 3D canonical points into the deformed space of each view and checking the means of the 2D object mask occupancies when projected:
\begin{equation}
    \Phi_o^c(\mathbf{x^c})= {{\sum\limits_{k=1}^{|\mathbf{G}|} r_k\mathcal{M}_k(\Pi_k(\mathcal{W}(\mathbf{x^c})))}\over{\sum\limits_{k=1}^{|\mathbf{G}|} r_{k}\mathcal{I}_k(\Pi_k(\mathcal{W}(\mathbf{x^c})))}}
\label{eq:whole_aggregation}
\end{equation}
where $\mathbf{G}$ is the set of generated images, $r_k$ is accumulation score for image $k$ based on predicted camera distribution, $\mathcal{M}_k$, $\mathcal{I}_k$ are each mask and image operator for $k$-th image which returns $1$ if the provided 2D value lies within mask/image space else $0$, and $\Pi_k$ is perspective projection for $k$-th image view.

\noindent \textbf{Uniform View Sampling.} Despite our viewpoint augmentation when prompting, the resulting images still may have biases toward specific viewpoints, making 3D reasoning difficult. Thus, we enforce uniform view sampling (similar to importance sampling) by dividing the azimuth region into a fixed number of bins and setting the accumulation score $r_k$ in Eq.~\ref{eq:whole_aggregation} as the inverse of camera numbers in the bin.

\noindent \textbf{Inference for Posed Space.} At inference, we can transform our occupancy probability distribution $\Phi_o^c(\mathbf{x}^c)$ defined in the canonical space into the pose-deformed space $\Phi_o(\mathbf{x}|\theta)$ by using backward skinning, unlike the forward skinning (refer to Eq.~\ref{eq:forwardkinematicpipeline}) during training. Specifically, we set $\Phi_o(\mathbf{x}|\theta)$ by warping $\mathbf{x}$ in pose-deformed space to $\mathbf{x}^c$ in canonical space and retrieving the learned occupancy probability $\Phi_o^c(\mathbf{x}^c)$. To this end, we compute the LBS weights in the \textit{pose-deformed} space (in contrast to training) and apply inverse transformation $\mathcal{W}^{-1}$.

\subsection{Selective Aggregation via Semantic Clustering}
\label{subsect:clustering}
The canonical distribution $\Phi_o^c(\mathbf{x}^c)$ from Eq.~\ref{eq:whole_aggregation} does not take into account the different semantics. However, the form of human-object interaction can differ even within the same object category, requiring corresponding variations of $\Phi_o^c(\mathbf{x}^c)$, as defined in Eq.~\ref{eq:HOI_distribtuion_canonical}.
To formulate this, we define the interaction type $s\in\mathbf{S}$ as a pair of prompt $p$ and body part $\mathbf{a}$ (optional) in contact with the object:
\begin{equation}
    \mathbf{S} = \{(p, \mathbf{a})~|~p \in \mathbf{P}, \mathbf{a} \in \mathbf{A}\}, 
    \label{eq:semantic_type}
\end{equation}
where $\mathbf{P}$ is the set of entire prompts produced and $\mathbf{A}$ is the set of body part segments, given as a part of SMPL mesh vertices. Intuitively, a prompt represents a semantic (e.g., ``playing with a ball''), and the body segment can further specify the interaction type (e.g., foot$\xrightarrow{}$``kicking'').

We compute $\Phi^c_o( \mathbf{x}^c | s)$ for each interaction type $s=(p,\mathbf{a})$ by aggregating a semantic cluster of images that depict such interactions, selected from $\mathbf{G}$. Specifically, we utilize images generated from the single prompt $p$. Body part $\mathbf{a}$ is used as a proximity cue to retain relevant image samples, where we consider the image irrelevant if the 3D rays from the object mask do not intersect with the interaction region of $\mathbf{a}$.
Our method of ``selective aggregation'' enhances spatial HOI reasoning for multimodal scenarios.
\section{Experiments}
\label{sec:experiments}
We provide both quantitative and qualitative comparisons to verify our method. 
In Sec.~\ref{subsec:pap}, we present a new metric, namely \textit{Projective Average Precision (PAP)}. 
We provide a brief explanation of PAP and detailed protocols in Supp. Mat.~\ref{appendix:protocolsforprojectiveaverageprecision}.
In Sec.~\ref{subsec:quantitative_evaluation}, we quantitatively compare our method with various ablations to provide justifications for our design choices. We also compare our method trained on synthesized images with the one trained on internet-crawled images, and show that synthesized images are more suitable for learning 3D HOI. In Sec.~\ref{subsec:qualitativeevaluation}, we show qualitative results. We demonstrate the quality of the learned HOI spatial distribution when various human pose is given for a diverse set of object categories. We then explore the effects of semantic types on the distribution by changing HOI prompts and body part specifications. We also analyze the effects of canonicalization via a comparison with an ablation. In Sec.~\ref{subsec:application}, we exemplify an application of our method to the downstream task: \textit{3D Human-Object Reconstruction from a Single-view Image}.
Refer to Supp. Mat.~\ref{appendix:dataset}$\sim$\ref{appendix:protocolsforprojectiveaverageprecision} for more details on each part, and Supp. Mat.~\ref{appendix:qualitativeevaluation} for extensive qualitative results.

\subsection{Dataset}
\label{subsec:dataset}
\noindent \textbf{Generated Dataset.} Since our method is a fully-autonomous and self-supervised approach, no external dataset is required for training. We test our method for 19 object categories from COCO~\cite{coco} (e.g., bicycle, chair) and 5 categories from LVIS~\cite{lvis} (e.g., hat, sweater), where we generate $5000\sim90000$ images per category.

\noindent \textbf{Image Search Dataset.} We provide a quantitative comparison between the results of our method using image search data and using synthesized images in Sec.~\ref{subsec:quantitative_evaluation}.
To this end, we use AutoCrawler~\cite{autocrawler} to collect images for category \textit{motorcycle} from the internet, using the same prompts used for preparing generated dataset.

\noindent \textbf{Extended COCO-EFT Dataset for Testing.} To quantitatively evaluate our learned HOI distributions, we use COCO dataset~\cite{coco}, which provides GT 2D object masks. In particular, we use COCO-EFT (val)~\cite{eft}, where pseudo GTs for 3D human poses are provided in SMPL~\cite{smpl} format. Among all samples in COCO-EFT, we perform a filtering process to keep only the samples where human-object interactions happen by retaining the samples with a single human and a single object with sufficient overlaps between them. After the filtering process, we compute perspective camera parameters similar to Sec.~\ref{subsec:pose_canonicalization} from the weak-perspective cameras provided in the dataset.

\subsection{Projective Average Precision}
\label{subsec:pap}
\begin{figure}[t]
    \centering
        \begin{subfigure}[b]{1.0\columnwidth}
             \centering
        \includegraphics[width=1.0\columnwidth]{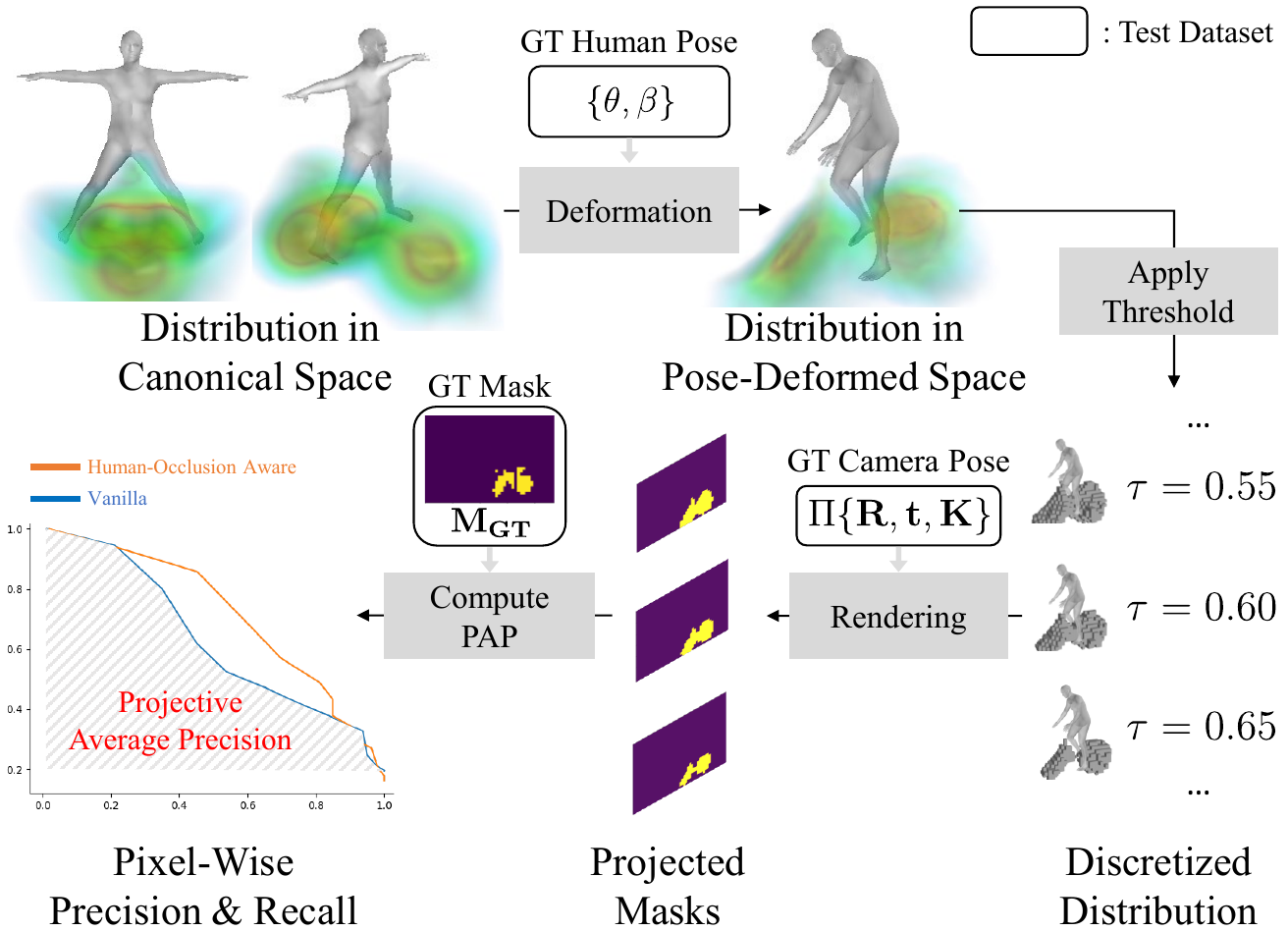}
        \caption{Overview.}
             \label{subfig:protocols}
        \end{subfigure}
        \begin{subfigure}[b]{1.0\columnwidth}
             \centering
        \includegraphics[width=1.0\columnwidth]{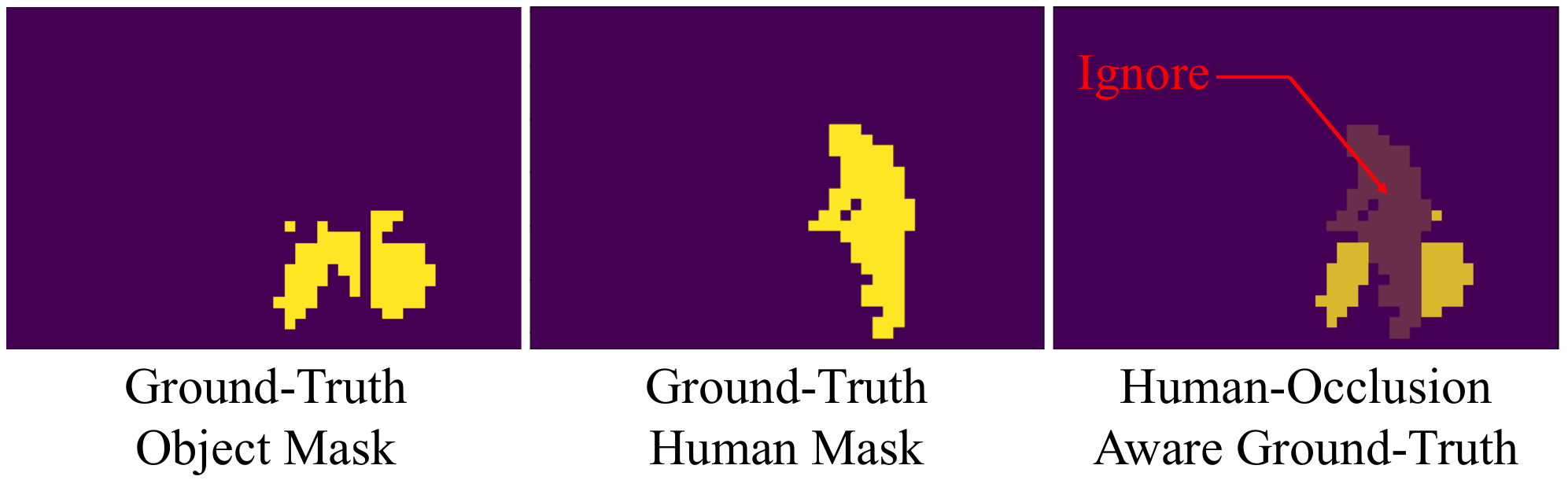}
        \caption{Human-occlusion aware mask.}
             \label{subfig:hoa_mask_comparison}
        \vspace{-13pt}
        \end{subfigure}
    \hfill
    \caption{\textbf{Overview of evaluation protocols for PAP.}}
    \label{fig:pap_protocol_viz}
    \vspace{-10pt}
\end{figure}

As no standard metrics exist for this task, we define a new evaluation metric: \textit{Projective Average Precision} (\textit{PAP}). Refer to Fig.~\ref{fig:pap_protocol_viz} for an overview of the PAP evaluation protocols.

\noindent \textbf{Evaluation Protocols.} We utilize the COCO-EFT~\cite{eft} dataset with pseudo-GT 3D human pose and object masks to compare the projection of estimated 3D distribution with GT object mask. Specifically, we first deform the canonical 3D distribution into pose-deformed space and apply threshold for discretization. We project discretized occupancy values to 2D using the perspective camera and compute \textit{pixel-wise} precision and recall between ground-truth object mask and projected occupancy. By setting multiple threshold values, we enable quantifying the validity of distribution regarding the intra-class object variation in terms of geometry. Note that we compute precision and recall for evaluation as our method aims to infer the \textit{object distribution}, not to reconstruct the exact geometry of the target.
The calculated precision and recall values are then employed to determine the average precision (AP) using two distinct methods (vanilla, strict). Averaging the AP values across all images within the category yields the \textit{Projective Average Precision (PAP)}. 
Further insights and comprehensive details can be found in Supp. Mat.~\ref{appendix:protocolsforprojectiveaverageprecision}.

\noindent  \textbf{Human-Occlusion-Aware PAP.} We observe that human masks frequently overlap with objects, potentially causing inaccuracies in evaluating rendered 3D occupancy against partially-removed ground-truth masks as shown in Fig.~\ref{subfig:hoa_mask_comparison}. To address this issue, we exclude precision and recall calculations within the 2D region of the human projection, considering the likelihood of occlusions.

\subsection{Quantitative Evaluation}
\label{subsec:quantitative_evaluation}
\noindent \textbf{Ablation Studies.} To justify our design choice, we quantitatively compare our method with various ablations.
We use mPAP (average of PAP for all categories) metric for comparison, and we report the results in Tab.~\ref{tab:quan-synthetic}. We list the baselines used for comparison and discussion for each result below. Unless specified, all baselines follow the same implementation details provided in Sec.~\ref{sec:method}.

\begin{itemize}[leftmargin=*]
    \vspace{-5pt}
    \itemsep 0em
    \item \textit{w/o Keypoint Filtering}: We deactivate keypoint filtering and conduct experiments using an equivalent number of images as in the case with keypoint filtering. We randomly sample from a combination of unfiltered and filtered images. We report a drop in performance when keypoint filters are not applied, which is attributed to erroneous 3D human predictions.

    \item \textit{w/o Canonicalization}: We compute the HOI distribution without pose canonicalization. Notably, we find that removing canonicalization damages the metrics by a large margin. This is an expected result, as removing canonicalization will disperse the 2D cues and limit the precise aggregation due to pose variation.

    \item \textit{10\%, 1\% Trained}: To see the effect of the number of generated data used for training (i.e., training time), we reduce the images by randomly selecting 10\% or 1\% of filtered images per prompt and run the 3D aggregation. The results imply that using more synthesized images for aggregation is beneficial for the quality of learned distribution, and the potential of our method for scalability.
\end{itemize}

\begin{table}[t]
    \caption{\textbf{Image Search vs. Synthesized Images: Statistics.}}
    \vspace{-5pt}
    \centering
    \resizebox{\columnwidth}{!}{
    \begin{tabular}{lcccc}
    \toprule
    Method & \# Images & \# Images after Filtering & Rejection-rate (\%)$\downarrow$ & Camera Entropy (bits)$\uparrow$ \\
    \midrule
    Image Search$_{\text{1k}}$ & 1201 & 391 & 67.44 & 10.04 \\
    Image Search$_{\text{10k}}$ & 9408 & 2526 & 73.15 & 12.25 \\
    Synthetic$_{\text{1k}}$ (Ours) & 1201 & 456 & 62.03 & 12.21 \\
    Synthetic$_{\text{10k}}$ (Ours) & 9408 & 3626 & \textbf{61.46} & \textbf{14.63} \\
    \bottomrule
    \end{tabular}
    }
    \label{tab:image-synth-dataset-stats}
    \vspace{-5pt}
\end{table}

\begin{table}[t]
    \caption{\textbf{Quantitative Evaluation Results.} (up) Results for ablation studies on COCO-EFT categories. (down) Results for quantitative comparison between image search and synthesize images on single category \textit{motorcycle}.}
    \label{tab:quan-synthetic}
    \small
    \vspace{-5pt}
    \centering
    \resizebox{\columnwidth}{!}{
        \begin{tabular}{lcccc}
        \toprule
        \multirow{2.5}{*}{Method}
        &\multicolumn{2}{c}{Vanilla}&\multicolumn{2}{c}{Human-Occlusion Aware} \\
        \cmidrule(lr){2-3}\cmidrule(lr){4-5}&mPAP$\uparrow$&mPAP$_\text{strict}$$\uparrow$ &mPAP$\uparrow$&mPAP$_\text{strict}$$\uparrow$ \\
        \midrule
        Ours$_\text{w/o Keypoint Filtering}$ & 18.31 & 15.90 & 19.97 & 16.52 \\
        Ours$_\text{w/o Canonicalization}$  & 17.52 & 15.12 & 19.56 & 15.85  \\
        Ours$_\text{1\%}$ & 14.98 & 13.81 & 15.21 & 13.73 \\
        Ours$_\text{10\%}$ & 17.90 & 16.13 & 18.86 & 15.82 \\
        Ours$_\text{full}$ & \textbf{19.86} & \textbf{17.18} & \textbf{20.28} & \textbf{16.80} \\
        \midrule
        Image Search$_{\text{1k}}$ &       53.64         & 52.15          & 60.08          & 56.86 \\
        Image Search$_{\text{10k}}$ &      55.97         & 54.59          & 63.62          & 61.11 \\
        Synthetic$_{\text{1k}}$ (Ours) &   56.33         & 55.11          & 63.91          & 61.85 \\
        Synthetic$_{\text{10k}}$ (Ours) &  \textbf{57.14}         & \textbf{55.19}          & \textbf{64.82}          & \textbf{62.11} \\
        \bottomrule
        \end{tabular}
    }
\vspace{-10pt}
\end{table}

\noindent \textbf{Image Search vs. Synthesized Images.} To validate the use of synthesized images, we compare our method's output for \textit{motorcycle} category across diverse datasets. We opt for \textit{motorcycle} due to the abundance of accessible and common images, which ensures a fair and rigorous comparison. We assess four settings, altering (1) data source (image search or synthesis) and (2) image count (1k or 10k). To ensure fairness, we randomly select images from the generated dataset, equivalent to the size of image search dataset. We follow the same procedures described in Sec.~\ref{sec:method} for filtering and aggregation. We report the dataset's retrieval statistics including post-filtering rejection rate, and camera distribution entropy for filtered images in Tab.~\ref{tab:image-synth-dataset-stats}. For the computation of camera distribution entropy, we fit Gaussian normals with a standard deviation of $\sigma=0.01$ to each camera's location on the unit sphere, using geodesic metrics.

We observe a higher and increasing rejection rate in the image search dataset compared to the stable, relatively low rate in the generated dataset. We attribute this to the word-occurrence-based search algorithm, which may retrieve images without HOI (e.g., commercials). Thus, we argue that the generated dataset offers enhanced scalability for HOI learning. Furthermore, the superior camera distribution entropy in synthesized images implies a richer diversity of camera poses than the image search dataset. Such diversity provides more informative content for 3D aggregation in the generated dataset. While our method addresses imbalanced camera distribution during aggregation through uniform view sampling (Sec.~\ref{subsec:pose_canonicalization}), results in Tab.~\ref{tab:quan-synthetic} consistently demonstrate the synthetic dataset's superior performance across metrics. This reaffirms the effectiveness of our data generation approach for HOI learning, even after minimizing information differences from camera imbalance.

\begin{figure}[t]
    \centering
    \includegraphics[width=\columnwidth]{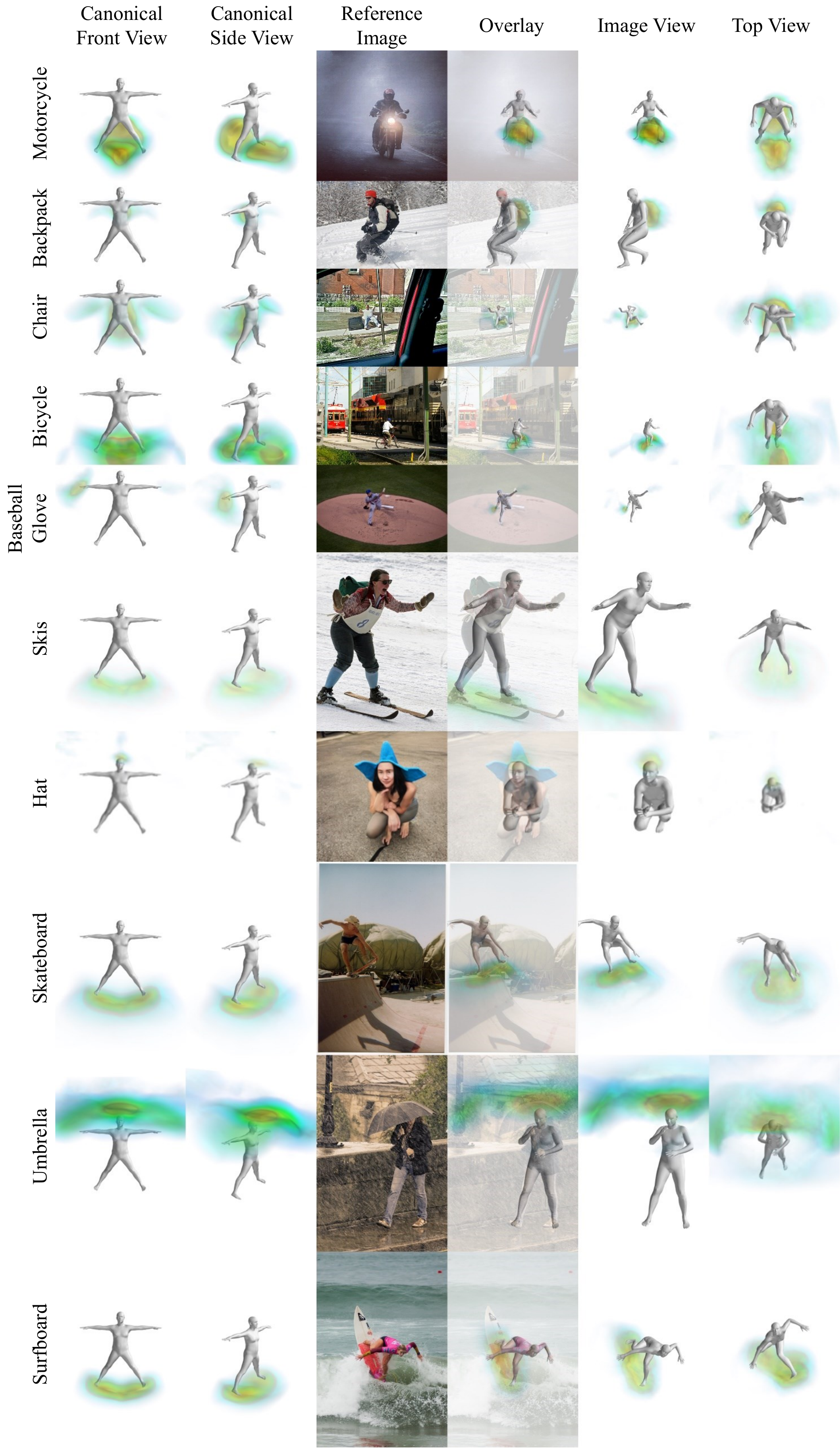}
    \caption{\textbf{Qualitative Results for Various Categories.}}
    \label{fig:qual_eval_various_categories}
    \vspace{-10pt}
\end{figure}

\begin{figure}[t]
    \centering
    \includegraphics[width=\columnwidth]{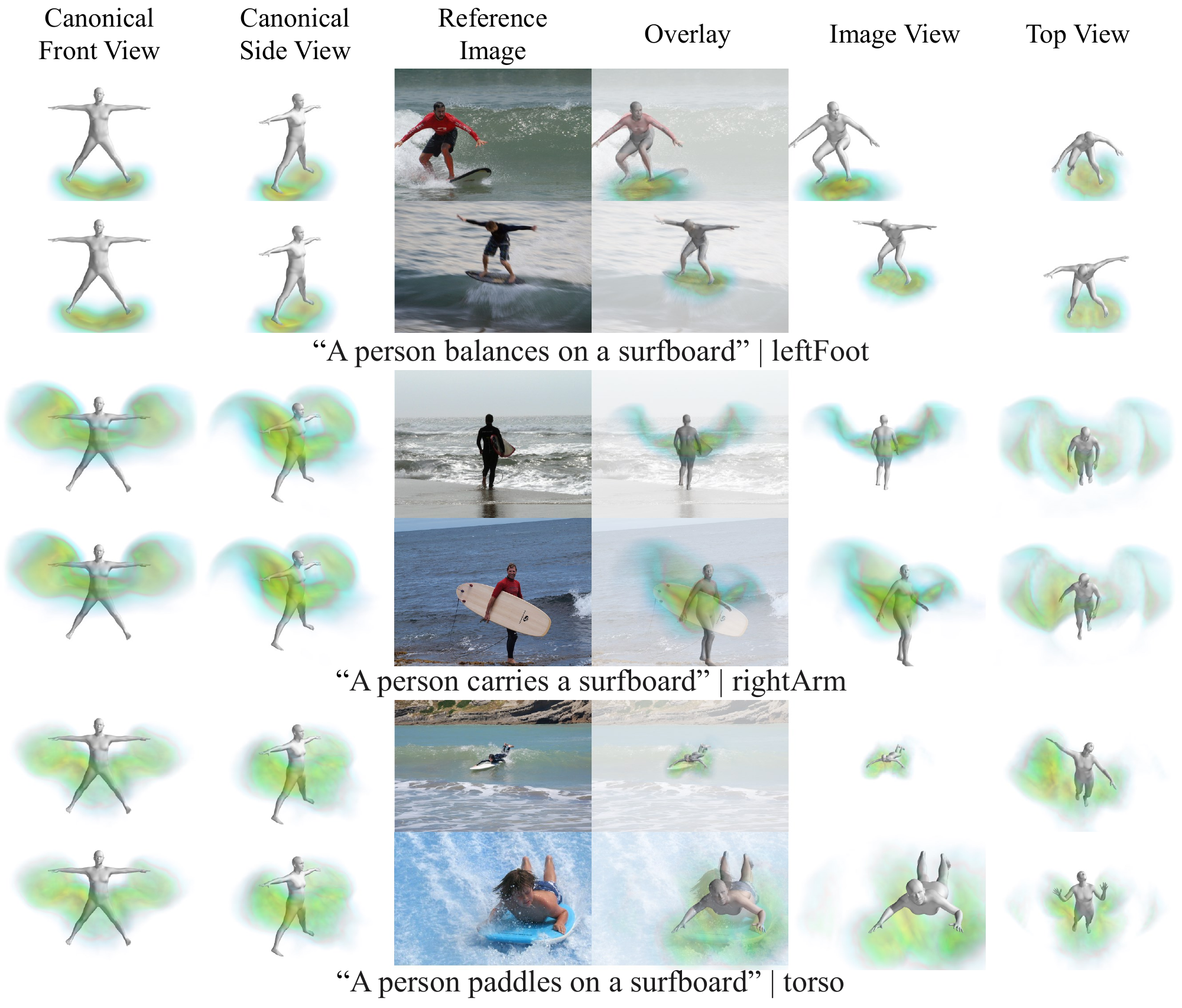}
    \caption{\textbf{Effects of semantic type condition on object distribution.} As semantic types vary, the canonical and pose-deformed object distribution changes accordingly. 
    }
    \label{fig:qual_eval_surfboard_diverse_semantics}
    \vspace{-15pt}
\end{figure}

\subsection{Qualitative Evaluation}
\label{subsec:qualitativeevaluation}
\noindent \textbf{Various Categories.} We demonstrate that our method can learn spatial human-object relationships for a diverse set of object categories in Fig.~\ref{fig:qual_eval_various_categories}. We use SMPL pose and reference image sampled from; the COCO-EFT~\cite{eft} dataset for COCO~\cite{coco} categories and generated dataset for LVIS~\cite{lvis} categories; to deform the distribution in canonical space to pose-deformed space. We visualize the semantic cluster that best matches the reference image. Refer to Supp. Mat.~\ref{appendix:qualitativeevaluation} for extensive results.

\begin{figure}[t]
    \centering
    \includegraphics[width=\columnwidth]{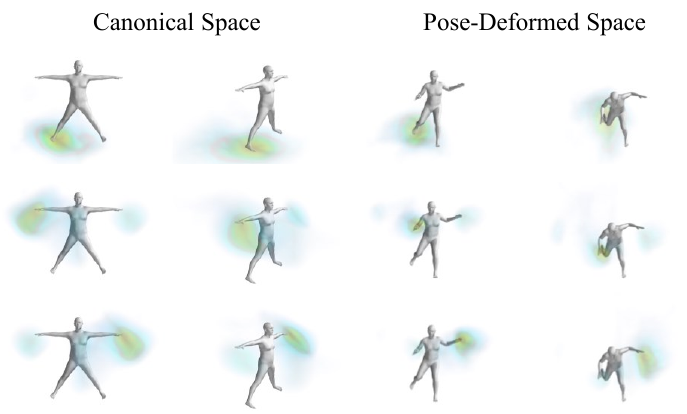}
    \caption{\textbf{Effects of body-part specification.} For category \textit{sports ball} and prompt ``A man playing with a sports ball'', body parts play a significant role as a condition for further specifying the HOI type.}
    \label{fig:qual_effect_body_part_sports_ball}
    \vspace{-5pt}
\end{figure}

\begin{figure}[t]
    \centering
    \includegraphics[width=\columnwidth]{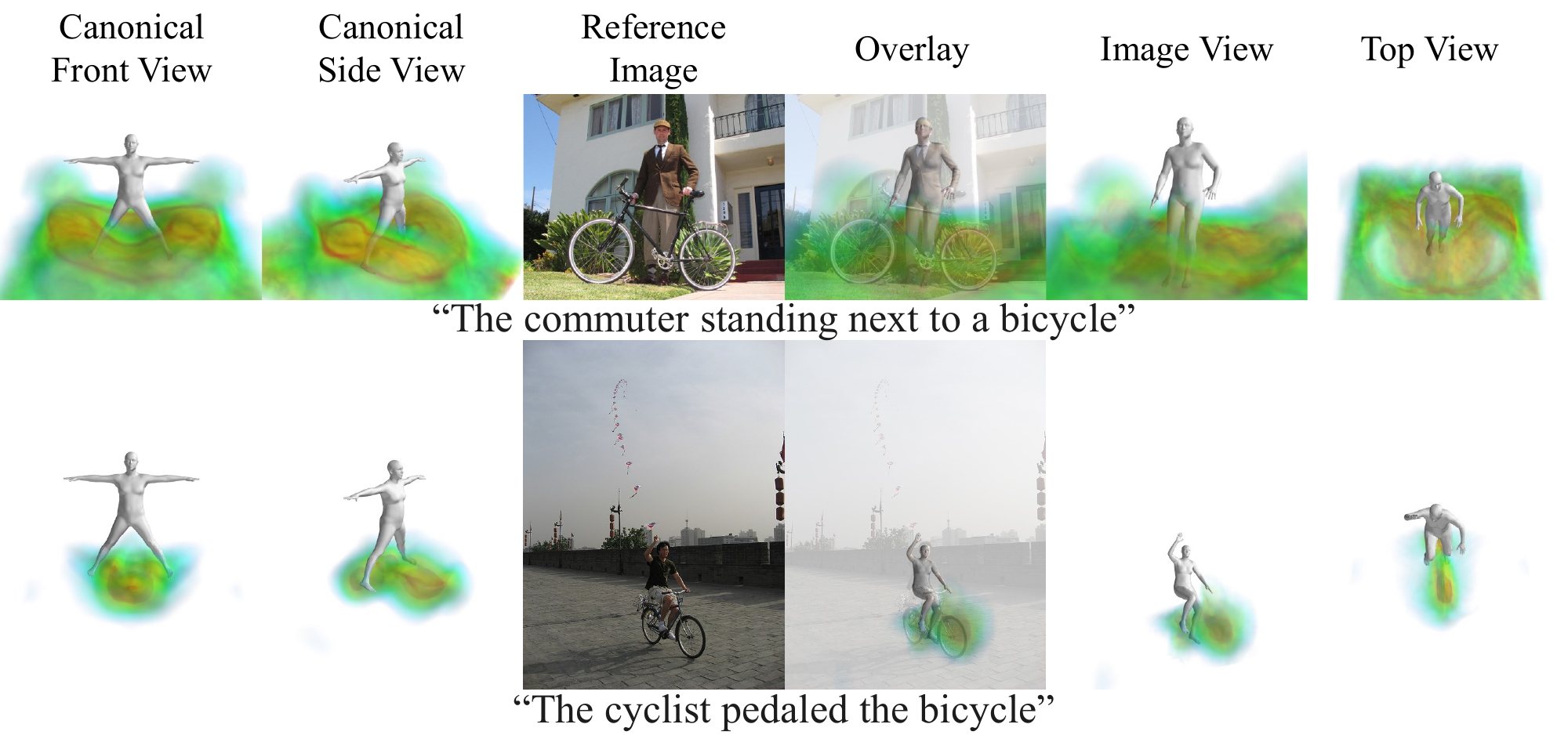}
    \caption{\textbf{Effects of prompt specification.} For category \textit{bicycle}, object distribution changes significantly as prompts representing the HOI type differs.}
    \label{fig:bicycle_various_semantics}
    \vspace{-5pt}
\end{figure}

\noindent \textbf{Effects of Semantic Type Condition.} We find that our set of learned distributions conveys the human-object interaction effectively. 
We report the examples of object distributions (category: \textit{surfboard}) in canonical and pose-deformed space for different semantic types $s=(p, \mathbf{a})$ in Fig.~\ref{fig:qual_eval_surfboard_diverse_semantics}. Notably, the object distribution of the surfboard for the given prompt ``A person carries a surfboard'' shows the mode near the human torso and arms: which is a plausible location for holding a surfboard when carrying around. We also report results when semantic body part $\mathbf{a}$ vary for the same semantic prompt description.
In Fig.~\ref{fig:qual_effect_body_part_sports_ball}, we observe that body parts act important as a condition when the provided semantic prompt contains little or less information on HOI; hence, ambiguous. For example, the semantics for ``A man playing with a sports ball'' can be different in terms of sports categories. In Fig.~\ref{fig:bicycle_various_semantics}, we can see the effects of prompt change when body parts are the same. While the semantic type with prompt ``The cyclist pedaled the bicycle'' return a shape similar to the original bicycle as expected, we observe the donut-shaped distribution as in Fig.~\ref{fig:bicycle_various_semantics} when we change the prompt into ``The commuter standing next to a bicycle''. We argue that such distribution conveys accurate semantics of spatial arrangements as humans can be oriented in arbitrary directions even if standing next to a bicycle, where we can assume such arrangement leads to rotationally-invariant distribution (e.g., donut-shape).

\begin{figure}[t]
    \centering
    \includegraphics[width=1.0\columnwidth]{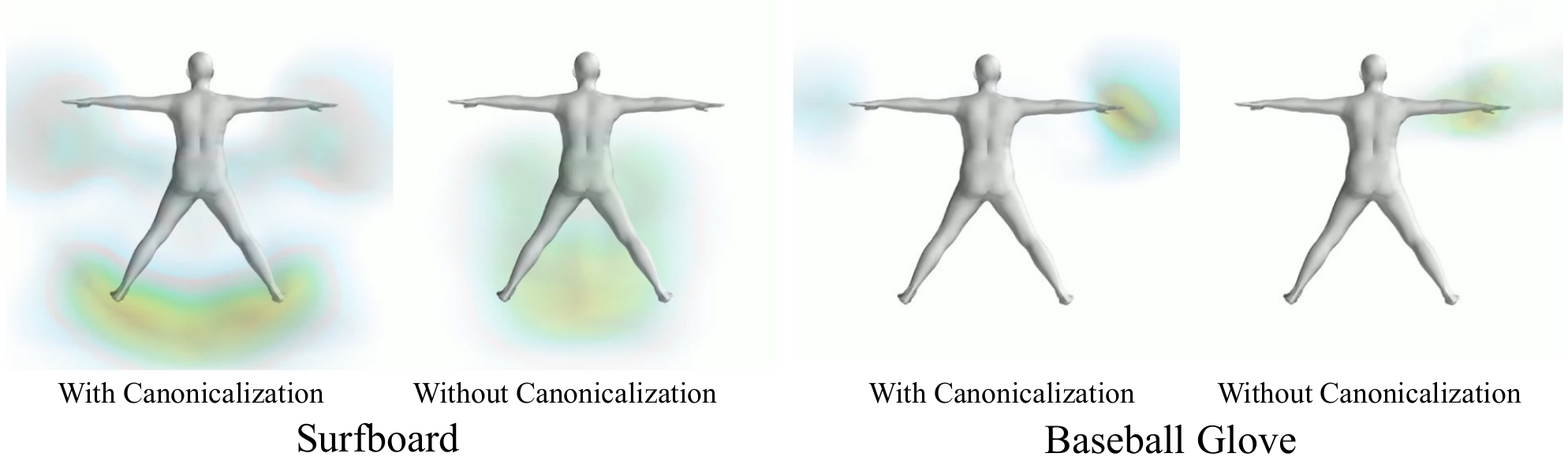}
    \caption{\textbf{Effects of canonicalization.} Removing canonicalization from our approach leads to scattered aggregation of 2D mask occupancies, resulting in blurry outcomes and diminished precision.}
    \label{fig:abl canon}
    \vspace{-10pt}
\end{figure}

\noindent \textbf{Effects of Canonicalization.} 
We ablate the effects of canonicalization and report results in Fig.~\ref{fig:abl canon}. From results, we can observe that precision is enhanced and variance is reduced when pose-canonicalization is applied, as the aggregation of 2D cues becomes more coherent compared to direct rays due to pose-variance.

\subsection{Application}
\label{subsec:application}
\begin{figure}[t]
    \centering
    \includegraphics[width=\columnwidth]{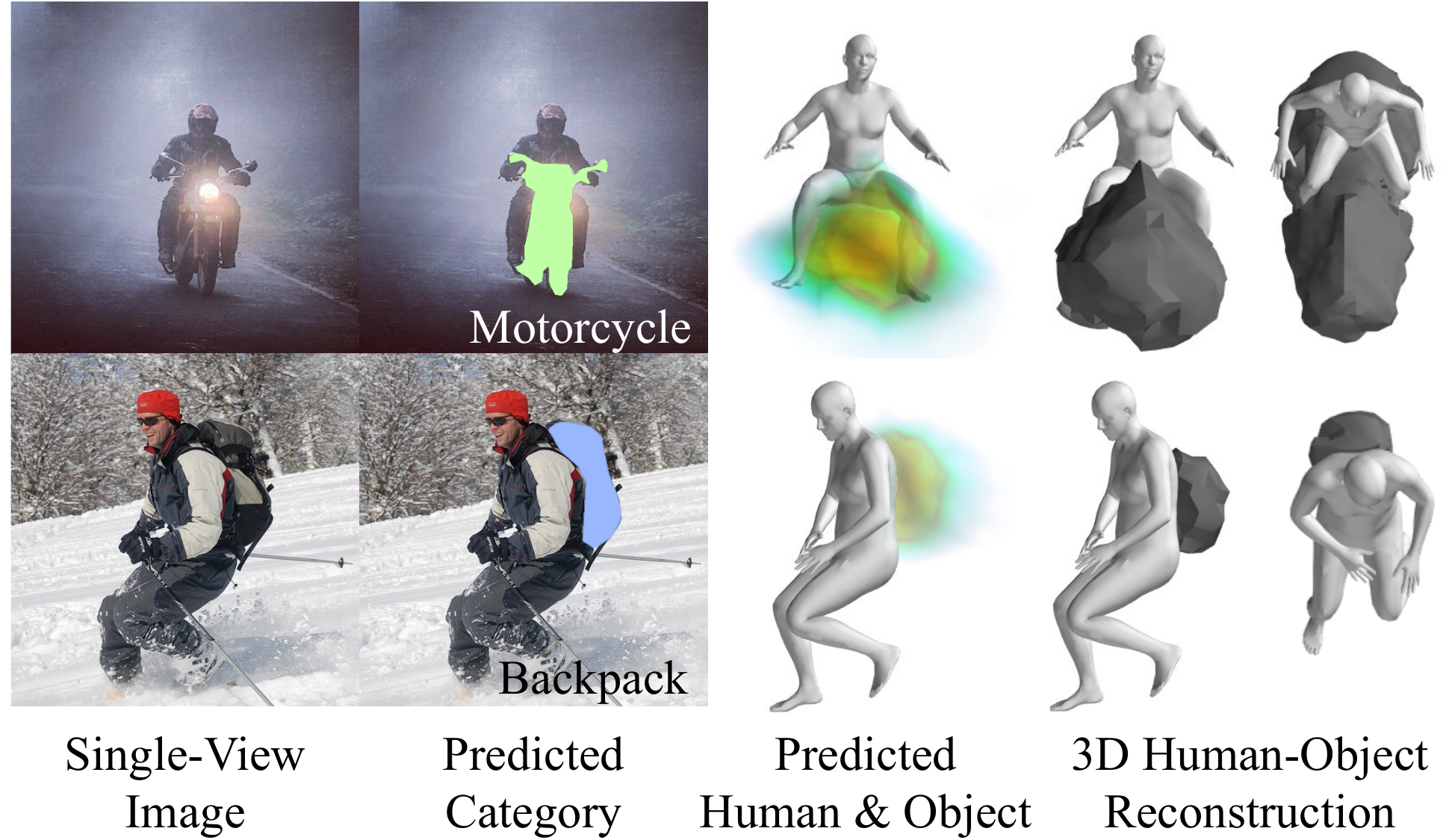}
    \caption{\textbf{3D Human-Object Reconstruction from a Single-view Image.} Using learned distribution as a prior, we can reconstruct 3D human and object from single image without any template mesh.}
    \label{fig:application}
    \vspace{-10pt}
\end{figure}

Our method returns generalizable 3D human-object spatial knowledge represented as probability occupancy distribution of the object, which can be easily applied to many downstream tasks. This section provides an example case of application to the following downstream task: \textit{3D Human-Object Reconstruction from a Single-view Image}. Given a single-view image, we apply off-the-shelf monocular 3D human pose estimator~\cite{frankmocap} to extract SMPL pose and off-the-shelf object detector~\cite{pointrend} to find the category of interacting object, where we find the object of interaction by computing bounding box overlap between human and object and picking the object with maximum value. Next, we extract 3D object occupancy distribution by simply inputting the category of the found object as a keyword to our method. Finally, we apply the marching cube algorithm~\cite{marchingcube} to the object distribution to convert 3D object occupancy into a 3D object mesh. Our method is applicable to in-the-wild images, fully automatic and template agnostic, in contrast to previous works that use template mesh to represent 3D objects~\cite {hoi_phosa, hoi_chore}. See Fig.~\ref{fig:application} for results.

\vspace{-5pt}
\section{Discussion}
\label{sec:conclusion}
Our method introduces a novel approach, enabling machines to probabilistically learn 3D human-object spatial relationships based on pose and HOI types. Unlike prior methods, ours extracts this information through self-supervision, eliminating the need for laborious annotations. We present multiple strategies, including leveraging a generative model for prompt/image synthesis to enhance image control, and a canonicalization-based framework for computing 3D spatial HOI fields from synthesized 2D images. Furthermore, we introduce a novel metric that demonstrates the high quality of our learned HOI distribution.
As the work presents a method to extract an intermediate representation, our approach offers numerous potential downstream applications; however, while possessing such potentials, the work is new and thus has limitations, such as low granularity and inaccurate modeling of distributions for small objects. We extensively discuss these limitations and potential future avenues in Supp. Mat.~\ref{appendix:limitations}.

\vspace{-15pt}
\blfootnote{
\noindent \textbf{Acknowledgements.}
We greatly appreciate Byungjun Kim for valuable insights and comments.
This work was supported by SNU-Naver Hyperscale AI Center, New Faculty Startup Fund from Seoul National University, SNU Creative-Pioneering Researchers Program, NRF grant funded by the Korea government (MSIT) (No. 2022R1A2C2092724), and IITP grant funded by the Korea government(MSIT) (No.2022-0-00156 and No.2021-0-01343). H. Joo is the corresponding author.
}

{\small
\setlength{\bibsep}{0pt}
\bibliographystyle{abbrvnat}
\bibliography{shortstrings, references}
}

\ifarxiv \clearpage \appendix
\label{appendix}

\twocolumn[
\begin{center}
    \Large\textbf{\paperTitle \\ (Supplementary Material)} 
\end{center}
]

\section{Additional Details for Method}
\label{appendix:additionaldetailsformethod}
In this section, we provide further details on the formulations and implementations regarding the method section (Sec.~\ref{sec:method}) of our main paper.

\subsection{Prompt Generation}
\label{appendix:promptgeneration}
We generate 3 to 20 different semantic HOI prompts via chatGPT~\cite{chatgpt} using the following query template, where \{\textbf{m}\} is replaced with an integer between 3$\sim$20 and \{\textbf{category}\} is replaced with one of our input category keywords:
\begin{quote}
    \textit{``Generate at most} $\{\textbf{m}\}$ \textit{simple subject-verb-object prompt where subject's} $\{$category is person/word is ``A person''$\}$ \textit{and object's category is} $\{\textbf{category}\}$. \textit{You should use diverse and general word but no pronoun for subject. Generated prompt must align with common sense. Verb must depict physical interaction between subject and object. Simple verb preferred.''}
\end{quote}     
The categories considered in this work are listed in Tab.~\ref{tab:full_category_stats}.
After generating HOI prompts, we augment each prompt with $m_{\text{aug}}=22$ different viewpoint augmentations, including:
\begin{quote}
    \textit{(no augmentation) / side view / back view / front view / top view / bottom view / realistic photo / seen from side / seen from back / seen from various views / seen from close view / seen from far away / scenic view / full body photo / hand photo / face photo / photo taken close to hand / photo taken close to face / selfie / close-up photo / hand only / face only}.
\end{quote}

\begin{table}[t]
    \caption{\textbf{Statistics for generated dataset for all categories.}}
    \centering
    \resizebox{1.0\columnwidth}{!}{
    \begin{tabular}{lcccc}
    \toprule
    Category & \# Prompts & \# Images & \# Images after Filtering & Rejection-rate (\%) \\
\midrule
Motorcycle & 12 & 47520 & 18228 & 61.64 \\
Bench & 12 & 31680 & 15193 & 52.04 \\
Backpack & 20 & 42240 & 7530 & 82.17 \\
Handbag & 4 & 23760 & 1374 & 94.22 \\
Tie & 10 & 31680 & 3358 & 89.40 \\
Frisbee & 3 & 21780 & 5502 & 74.74 \\
Skis & 12 & 34848 & 14547 & 58.26 \\
Snowboard & 4 & 7920 & 3251 & 58.95 \\
Sports ball & 8 & 19008 & 1611 & 91.52  \\
Baseball glove & 5 & 11880 & 260 & 97.81 \\
Skateboard & 16 & 50688 & 12069 & 76.19 \\
Surfboard & 16 & 73920 & 25352 & 65.70 \\
Tennis racket & 15 & 47520 & 19827 & 58.28 \\
Cell phone & 20 & 84480 & 2344 & 97.23 \\
Bicycle & 18 & 42768 & 11361 & 73.44 \\
Umbrella & 7 & 16632 & 2720 & 83.65 \\
Chair & 17 & 47124 & 10185 & 78.39 \\
Bed & 5 & 11880 & 1856 & 84.38 \\
Laptop & 17 & 53856 & 1890 & 96.49 \\
Hat$^*$ & 10 & 31680 & 173 & 99.45 \\
Sweater$^*$ & 3 & 5940 & 689 & 88.40 \\
Sunglasses$^*$ & 3 & 5940 & 78 & 98.69 \\
Soccer ball$^*$ & 5 & 15840 & 1976 & 87.53 \\
Scarf$^*$ & 5 & 15840 & 461 & 97.09 \\
\bottomrule
\multicolumn{4}{l}{* denotes LVIS~\cite{lvis} categories} \\
\end{tabular}
\label{tab:full_category_stats}
}
\vspace{-15pt}
\end{table}

\subsection{Synthesizing Text-Conditioned Images via Diffusion}
\label{appendix:synthesizingtextconditionedimagesviadiffusion}
We use publicly available Stable-Diffusion~\cite{diffusion_ldm} model (version: 1.4) for text-to-image synthesis. We generate $512\times512$ images upsampled from the generated 8$\times$ downsampled latents.  We sample images with $50$ denoising steps using Pseudo-Numerical-Sampling methods~\cite{diffusion_pseudonumerical} with classifier-free guidance scale of $7.5$.

\subsection{Filtering}
\label{appendix:filtering}
We apply a cascaded framework to remove unrelated images. We use PointRend~\cite{pointrend} for bounding-box detection and instance segmentation of COCO~\cite{coco} categories. For LVIS~\cite{lvis} categories (e.g., hat, sunglasses, sweater), we use publicly available pretrained Mask-RCNN~\cite{maskrcnn} model from detectron2~\cite{detectron2} where we set the segmentation threshold as $0.8$. We post-process the predicted instances by removing duplicated bounding boxes on the same target object.
Specifically, we remove the lower-confidence instance if two bounding boxes of the same category overlap with the \emph{intersection over smaller bounding box} value bigger than $0.8$, with exceptions of bounding boxes with confidence over $0.98$.
We also filter the images with multiple or none human/object (target category) instances. We then reject images when the \emph{intersection over object bounding box} value between the human box and object box is smaller than $0.1$, assuming there is no interaction between them.
For keypoint-filtering, we represent the keypoints in COCO~\cite{coco} format. We use a top-down approach using publicly available pretrained HRNet~\cite{hrnet}+Darkpose~\cite{darkpose} (provided by MMPose~\cite{mmpose2020}) with a confidence threshold of $0.7$ for keypoint prediction. We exclude the image if no shoulder joints (i.e., \textit{left-shoulder, right-shoulder}) or no hip joints (i.e., \textit{left-hip, right-hip}) exist. Finally, we reject the images with a very small human bounding box that returns no prediction from 3D human pose estimator~\cite{frankmocap}.

\subsection{Viewpoint Estimation via 3D Human Pose Estimation}
\label{appendix:viewpointestimationvia3dhumanposeestimation}
While the weak perspective is sufficient for projection, we employ a perspective camera to consider the distance between the camera and the human subject.
To convert the weak perspective camera model $\mathbf{\pi}$ and camera-centric orientation $\mathbf{\phi}$ (refer to Eq.~\ref{eq:pose3d}) into a perspective camera model $\Pi$ in the person-centric coordinate system (i.e., the origin is defined at the pelvis), we apply optimization to compute $\Pi$ by aligning 3D joint projections and $\mathbf{j}$:
\begin{equation}
    \Pi^{*} = \argmin_{\Pi} \sum_i || \Pi(\mathbf{J}^{0}_{i}) - \mathbf{j}_{i} ||^2,
\label{eq:ortho2persp}
\end{equation}
where $i$ is the joint index and $\mathbf{J}^{0}$ are the 3D joints of the SMPL model in person-centric coordinate, which is simply obtained by putting zero orientation for $\mathbf{\phi}$ while keeping other parameters $\mathbf{\theta}$ and $\mathbf{\beta}$. 
We include the joints $\mathbf{j}_i$ outside the image range for optimization described in Eq.~\ref{eq:ortho2persp}. We optimize the joint-reprojection loss using Adam~\cite{adam} with a learning rate of $0.01$ for $2400$ iterations. We early-terminate if the joint-reprojection loss is below 0.7 in pixel scale. We initialize camera intrinsic parameters with field-of-view $46.4^\circ$ and camera extrinsic parameters with the rotation matrix achieved by Rodrigues formula using $\phi$ and the translation vector by $[t_x,t_y,2f/s]$ where $t_x,t_y,s$ are each $x$,$y$-direction translations, and scale from weak-perspective camera $\pi$.

\begin{figure}[t]
    \centering
    \includegraphics[width=\columnwidth]{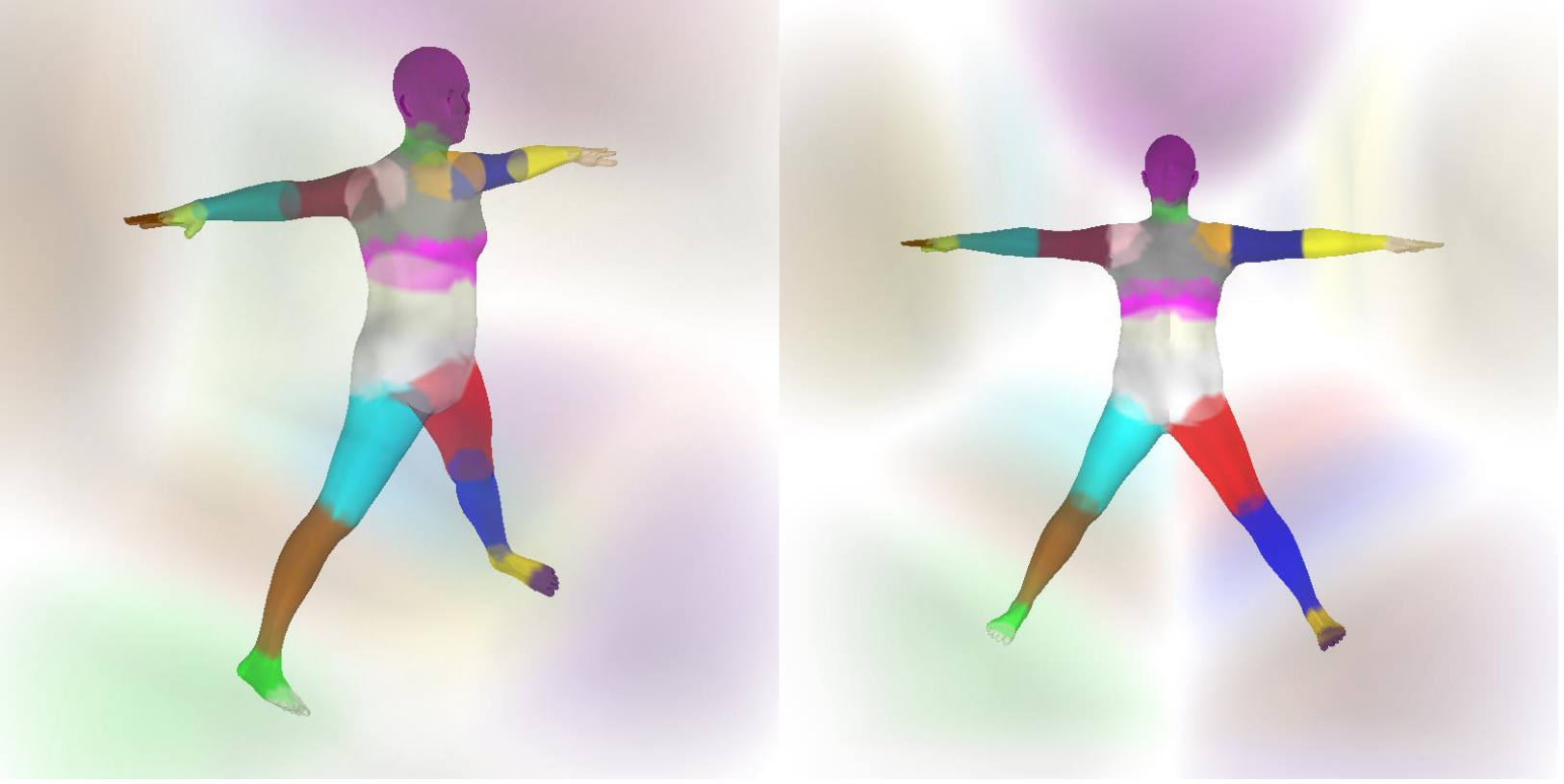}
    \caption{\textbf{Visualization of LBS weights.}}
    \label{fig:lbs_weights_visualized_supplemenetary}
\end{figure}

\subsection{3D Occupancy Estimation via Human Pose Canonicalization}
\label{appendix:3doccupancyestimationviahumanposecanonicalization}
We define LBS weights for arbitrary point $\mathbf{x}^c$ in canonical space following Eq.~\ref{eq:lbs_nearest_neighbor} with $\mathbf{k}=30$. We encourage ``zero motions'' by mixing computed LBS weights with standard basis vector for 1st dimension $\mathbf{e}_1\in\mathbb{R}^{24}$:
\begin{equation}
    \label{eq:deweighting}
    \omega_{\text{deweight}}(\mathbf{x}^c) = (1-\alpha)\mathbf{e}_1 + \alpha \omega(\mathbf{x}^c)
\end{equation}
\begin{equation}    
    \alpha=|{\max({{\tau-d_{\text{min}}}\over{\tau+d_{\text{min}}}}, 0)}|^s
\end{equation}
where $\tau$ and $s$ are each bandwidth and smoothing hyperparameters we choose as $\tau=0.8, s=0.25$, and $d_{\text{min}}$ is the distance between $\mathbf{x}^c$ and nearest SMPL mesh vertex. Note that the 1st element of LBS weight is related to the pelvis joint, which is fixed to the origin and aligned in a person-centric coordinate system as we set $\phi=0$; hence, the transformation matrix is identity in $\text{SE}(3)$. Finally, we apply Laplace smoothing over the entire grid 30 times to smooth LBS weights, similar to SelfRecon~\cite{selfrecon}. See Fig.~\ref{fig:lbs_weights_visualized_supplemenetary} for visualization of extended LBS weights.

\subsection{Uniform View Sampling}
\label{appendix:uniformviewsampling}
Before performing aggregation, we first check the camera distribution and assign the accumulation score $r_k$ for each image. Specifically, we divide the azimuth region $[0,2\pi)$ into 12 equispaced bins (each $\pi\over 6$ long) and count the number of cameras in each bin. If the camera associated with the image falls into a specific bin, $r_k$ is set as the inverse of the camera numbers in that bin. After assigning $r_k$ for all images, we then perform the aggregation following Eq.~\ref{eq:whole_aggregation} or Eq.~\ref{eq:whole_aggregation_with_semantic_clustering}.

\subsection{Inference for Posed Space}
\label{appendix:inferenceforposedspace}
At inference, we first calculate LBS weights that transform the voxels from pose-deformed space to canonical space in the same way of Eq.~\ref{eq:lbs_nearest_neighbor} and Eq.~\ref{eq:deweighting}, where in this case $\mathbf{x}^c$ is replaced with $\mathbf{x}$ and $\mathbf{v}_i$ corresponds to location of $i$-th vertex in SMPL in pose-deformed space. Denoting the $j$-th LBS weights in pose-deformed space as $\omega^{\text{inv}}_j(\mathbf{x}; \theta)$, we inversely deform the voxels in pose-deformed space as:
\begin{equation}
    \mathbf{x}^c = \mathcal{W}^{-1}(\mathbf{x}) = \sum\limits_{j=1}^{n_b} \omega_j^{\text{inv}}(\mathbf{x}; \theta)\cdot \mathbf{B}_j(\theta_j)^{-1}\cdot \mathbf{x}
\end{equation}
We set $\Phi_o(\mathbf{x}|\theta)$ as the learned occupancy $\Phi_o^c(\mathbf{x}^c)$ in canonical distribution to infer pose-deformed distribution.

\begin{table}[t]
    \tiny
    \caption{\textbf{Definition of Body Parts.} Each body part is defined by merging multiple SMPL body segmentation maps from Meshcapade~\cite{meshcapade}.
    }
    \centering
    \resizebox{0.8\columnwidth}{!}{
    \begin{tabular}{rl}
    \toprule
    Body Part & SMPL Body Segmentation Labels \\
\midrule
    rightHand & rightHand, rightHandIndex1 \\
    leftHand & leftHand, leftHandIndex1 \\
    rightArm & rightArm, rightForeArm \\
    leftArm & leftArm, leftForeArm \\
    rightLowerLeg & rightLeg \\
    leftLowerLeg & leftLeg \\
    rightUpperLeg & rightUpLeg \\
    leftUpperLeg & leftUpLeg \\
    rightFoot & rightFoot, rightToeBase \\
    leftFoot & leftFoot, leftToeBase \\
    torso & spine, spine1, spine2, leftShoulder, rightShoulder \\
    face & head, neck \\
\bottomrule
\end{tabular}
\label{tab:smplmeshmapping}
}
\end{table}

\subsection{Selective Aggregation via Semantic Clustering}
\label{appendix:selectiveaggregationviasemanticclustering}
We note that body part $\mathbf{a}\in\mathbf{A}$ is a hyperparameter that can be easily given by annotating a set of SMPL mesh vertices with a body part label. In practice, we define $\mathbf{A}$ as 12 body parts obtained by merging publicly available SMPL segmentation maps from Meshcapade~\cite{meshcapade}. 
Correspondence between defined body parts and SMPL body segmentation labels is provided in Tab.~\ref{tab:smplmeshmapping}.
Representing each body part $\mathbf{a}\in\mathbf{A}$ as a binary operator that outputs $1$ if the corresponding SMPL mesh vertex is part of the body part $\mathbf{a}$ else $0$, we can define the interaction region for $\mathbf{a}$ in the canonical space as below:
\begin{equation}
    \mathbf{I}_\mathbf{a} = \bigcup\limits_{i;~\mathbf{a}(\mathbf{v}^i)=1}\{\mathbf{x}^c|\ ||\mathbf{x^c}-\mathbf{v}_i||\leq\epsilon\}    
\end{equation}
where $\epsilon$ is the interaction threshold, where we set as $\epsilon=0.13$.
The interaction region $\mathbf{I}_{\mathbf{a}}$ plays a role of determining whether the provided image contains the HOI involving contact with body part $\mathbf{a}$. Specifically, we ignore the image if none of the 3D canonical points within $\mathbf{I}_\mathbf{a}$ are warped and projected into the object mask. Putting it all together, selective aggregation for 3D occupancy can be described as:
\begin{equation}
    \label{eq:whole_aggregation_with_semantic_clustering}
    \resizebox{1.0\columnwidth}{!}{$
    \Phi_o^c(\mathbf{x^c}|s)= {{\sum\limits_{k=1}^{|\mathbf{G}(p)|} r_k\mathcal{M}_k(\Pi_k(\mathcal{W}(\mathbf{x^c})))\cdot\mathbf{1}(1\in\mathcal{M}_k(\Pi_k(\mathcal{W}(\mathbf{I}_\mathbf{a}))))}\over{\sum\limits_{k=1}^{|\mathbf{G}(p)|} r_{k}\mathcal{I}_k(\Pi_k(\mathcal{W}(\mathbf{x^c})))\cdot\mathbf{1}(1\in\mathcal{M}_k(\Pi_k(\mathcal{W}(\mathbf{I}_\mathbf{a}))))}}
    $}
\end{equation}
where $\mathbf{G}(p)$ denotes the set of generated images from single HOI prompt $p\in\mathbf{P}$ and $\mathbf{1}(\cdot)$ is a binary operator that returns $1$ if the provided input is true else $0$. At inference, learned occupancy probabilities are used to compute $\Phi_o(\mathbf{x}|\theta,s)$ (refer to Eq.~\ref{eq:HOI_distribtuion}) as described in Supp. Mat.~\ref{appendix:inferenceforposedspace}.

\section{Additional Details for Experiments}
\label{appendix:additionaldetailsforexperiments}
\subsection{Dataset}
\label{appendix:dataset}
\noindent \textbf{Generated Dataset.} We refer readers to Tab.~\ref{tab:full_category_stats} for full per-category statistics for the generated dataset.

\noindent \textbf{Image Search Dataset.} We use $m=12$ prompts (same as the prompts used for generating dataset) for category \textit{motorcycle} and $m_{\text{aug}}=22$ viewpoint augmentations (same as in Supp. Mat.~\ref{appendix:promptgeneration}) for image search, and we set the desired number of retrieved images as $N=1000$ or $N=10000$. We crawl up to $\lfloor \tau_{\text{mult}} \times  {N\over m \times m_{\text{aug}}} \rfloor + \tau_{\text{add}}$ image links, where $\tau_{\text{mult}}$ and $\tau_{\text{add}}$ are introduced to tolerate the number of undownloadable/unreadable files. Specifically, we set $\tau_{\text{mult}}=1.1$ and $\tau_{\text{add}}=1$. Subsequently, collected images are resized to a shorter side length of 512 and center-cropped, resulting in $512\times512$ images.

\noindent \textbf{Extended COCO-EFT Dataset for Testing.} We provide detailed dataset preparation procedures for the COCO-EFT~\cite{eft} dataset. Similar to filtering (Sec.~\ref{subsec:synthesize_images}), we only retain samples with a single human, single object (of target category), and filter the images based on \emph{intersection over smaller bounding box} value between the human and the object. We assume no human-object interaction if this value is below 0.5. It is important to note that we do not apply any manual filtering to ensure fairness in our evaluation. After filtering, we compute the perspective camera parameters following Supp. Mat.~\ref{appendix:viewpointestimationvia3dhumanposeestimation}, except we optimize for 3000 iterations and early-terminate if the joint-reprojection loss is below 0.5 in pixel scale. We reject the image if the joint-reprojection loss is over 1.0 in pixel scale. To ensure multi-viewpoint evaluation, we only use categories with more than 30 images in the extended COCO-EFT dataset. Refer to Tab.~\ref{tab:test_dataset_stats} for the summary of statistics.

\begin{table}[t]
    \tiny
    \caption{\textbf{Statistics in the Extended COCO-EFT Dataset.} We report number of images for categories with a minimum of 30 images in COCO-EFT~\cite{eft} dataset.}
    \centering
    \resizebox{0.8\columnwidth}{!}{
    \begin{tabular}{lc}
    \toprule
    Category & Number of Images in Dataset \\
\midrule
Motorcycle & 36 \\
Bench & 37 \\
Backpack & 83 \\
Handbag & 47 \\
Tie & 37 \\
Frisbee & 36 \\
Skis & 86 \\
Snowboard & 67 \\
Sports ball & 67 \\
Baseball glove & 49 \\
Skateboard & 176 \\
Surfboard & 110 \\
Tennis racket & 117 \\
Cell phone & 60 \\
\bottomrule
\end{tabular}
\label{tab:test_dataset_stats}
}
\end{table}

\subsection{Projective Average Precision}
\label{appendix:protocolsforprojectiveaverageprecision}
We provide detailed protocols for PAP evaluation and intuition behind each step in this section. Briefly speaking, the PAP metric quantifies the validity of object occupancy distribution in pose-deformed 3D space (current space) without 3D annotations by comparing the projection of distribution with 2D annotation from multiple viewpoints.
Given the category keyword to evaluate, we first start by deforming the distribution from canonical space to pose-deformed space using annotated SMPL pose from the test dataset following the inference method described in Sec.~\ref{subsec:pose_canonicalization}. Note that we can get probability occupancy values for equispaced gridpoints in pose-deformed space. We discretize the distribution in pose-deformed space for various threshold values and project binary occupancy using an annotated perspective camera. Discretization is applied to bypass the ambiguity of mixing probabilities when more than one 3D probability value falls into the same pixel in 2D. We use all thresholds from $0.01 \sim 1$ equispaced with interval $0.01$. Using multiple thresholds enables us to evaluate the distribution regarding the intra-class variation of object geometry. Next, we compute \textit{pixel-wise} precision and recall between rendered mask and annotated object segmentation mask for all thresholds. Specifically, we downsample rendered mask and object segmentation mask preserving aspect ratio with a shorter length being 32 before we compute precision and recall. We downsample for two reasons: (1) to allow other 3D representations that require high-compute in the PAP evaluation pipeline, and (2) to tolerate the variance in object size and geometry. Note that the goal of the PAP metric is to evaluate \textit{validity} of the distribution, not the accuracy or quality of the reconstruction. Finally, we compute interpolated AP similar to Pascal VOC 2008~\cite{pascal-voc-2008} using precision-recall values, which are subsequently averaged across all test images within the given category to yield the PAP value. We report two different PAP metrics in terms of interpolation methods when predicted occupancy is entirely $0$ after discretization:
\begin{itemize}
    \item \textit{Vanilla}: Uses the highest precision value from lower thresholds when discretized distribution is entirely $0$
    \item \textit{Strict}: Sets precision value as $0$ when discretized distribution is entirely $0$
\end{itemize}
Note that Human-Occlusion-Aware PAP metrics follow the same protocols, except we exclude the human-occluded region when computing precision and recall.
We do not apply semantic clustering during quantitative evaluation, i.e., we evaluate marginalized distribution aggregated with all images generated from all prompts per category.

\section{Additional Qualitative Results} 
\label{appendix:qualitativeevaluation}
We report additional qualitative results for various categories in this section. Same as in Sec.~\ref{subsec:qualitativeevaluation}, we use SMPL pose sampled from the extended COCO-EFT dataset for COCO categories or generated dataset for LVIS categories to deform the distribution in canonical space to pose-deformed space. See Fig.~\ref{fig:qual backpack}~$\sim$~\ref{fig:qual tie} for results.

\section{Limitations \& Future Works}
\label{appendix:limitations}
\noindent \textbf{Granularity.} Our method returns a plausible set of object distributions; however, we represent them as a low-resolution voxel field (resolution $48^3$), which limits the representability and granularity of the results. Future research can explore alternative 3D representations (e.g., volume rendering-based methods similar to NeRF~\cite{nerf}) to improve computation efficiency and achieve higher quality.

\noindent \textbf{Problems with Small Objects.}
Our method particularly shows weakness in small objects, especially those interacting with hands, primarily due to heavy occlusion and expressivity constraints in the SMPL model. For future research, employing the SMPL-X~\cite{smplx} representation to learn distributions for small objects interacting with hands, or using close-view cameras from various angles to reduce occlusion, could be beneficial.

\noindent \textbf{Bias and Artifacts in Synthesized Images.}
We use viewpoint augmentation to control the camera distribution during image synthesis; however, this method lacks full controllability and requires improvement. Although we minimize this effect by assigning camera distribution-aware accumulation scores, there is still a possibility of bias. One potential approach to address this challenge is by employing the PerpNeg algorithm~\cite{perpneg} to enhance viewpoint control during generation.
Additionally, synthesized images are likely to contain artifacts, which could propagate errors in later steps (e.g., human prediction) and lead to incorrect modeling of the occupancy probability distribution. Improving image synthesis methods will help mitigate such challenges.

\noindent \textbf{Bias due to Heavy Filtering.}
As our filtering strategy involves various off-the-shelf methods, employing heavy filtering may introduce bias. For example, the object detection method we use may be imperfect and could filter out images even if an object is present, leading to bias in the generated dataset after filtering. Consequently, this may result in bias in the occupancy probability distribution. Soft filtering methods (i.e., applying confidence weights to each image instead of removing images with hard thresholds) may be an alternative, which we leave for future work.

\noindent \textbf{Modelling Multimodal Scenarios.} The semantic clustering step in our method provides a means to understand objects that humans can interact with in various ways. While our method effectively models plausible HOI exhibiting specific interaction types, it requires manual definition of body parts and specification of HOI prompts to represent the semantics. Additionally, relying on user evaluation for identifying plausible semantic clusters hinders the efficient and automatic expansion of the corpus, as this process becomes manual. We acknowledge the need for further research to enhance the automatic generation of 3D HOI spatial relations without this manual constraint.

\noindent \textbf{Category Limits.}
Currently, our method mainly considers categories from COCO~\cite{coco} and a few categories from LVIS~\cite{lvis} due to the availability of the object detection and segmentation method. A promising future direction is to replace this current object detector with open-vocabulary models like ODISE~\cite{odise} to incorporate additional categories.

\noindent \textbf{Evaluation Metrics.}
Our evaluation metric (PAP) has room for improvement. For example, our current method utilizes a simple downsampling strategy to smooth the distribution during the protocol, which could be enhanced with other strategies (e.g., kernel-smoothing methods). 
Additionally, it is worth exploring enhancements to the metric that can effectively quantify the validity of multimodal distributions.

\noindent \textbf{Potential Downstream Applications.}
Our method exhibits myriads of potential downstream applications, such as; (1) using our extracted knowledge as prior for HOI modeling (e.g., replacing interaction labels in PHOSA~\cite{hoi_phosa}); (2) improving action recognition methods based on the current pose of the human and object; (3) scene generation and object localization from human postures; or (4) applications for robotics.

\newpage\begin{figure*}[p]
    \centering
    \includegraphics[width=2.0\columnwidth]{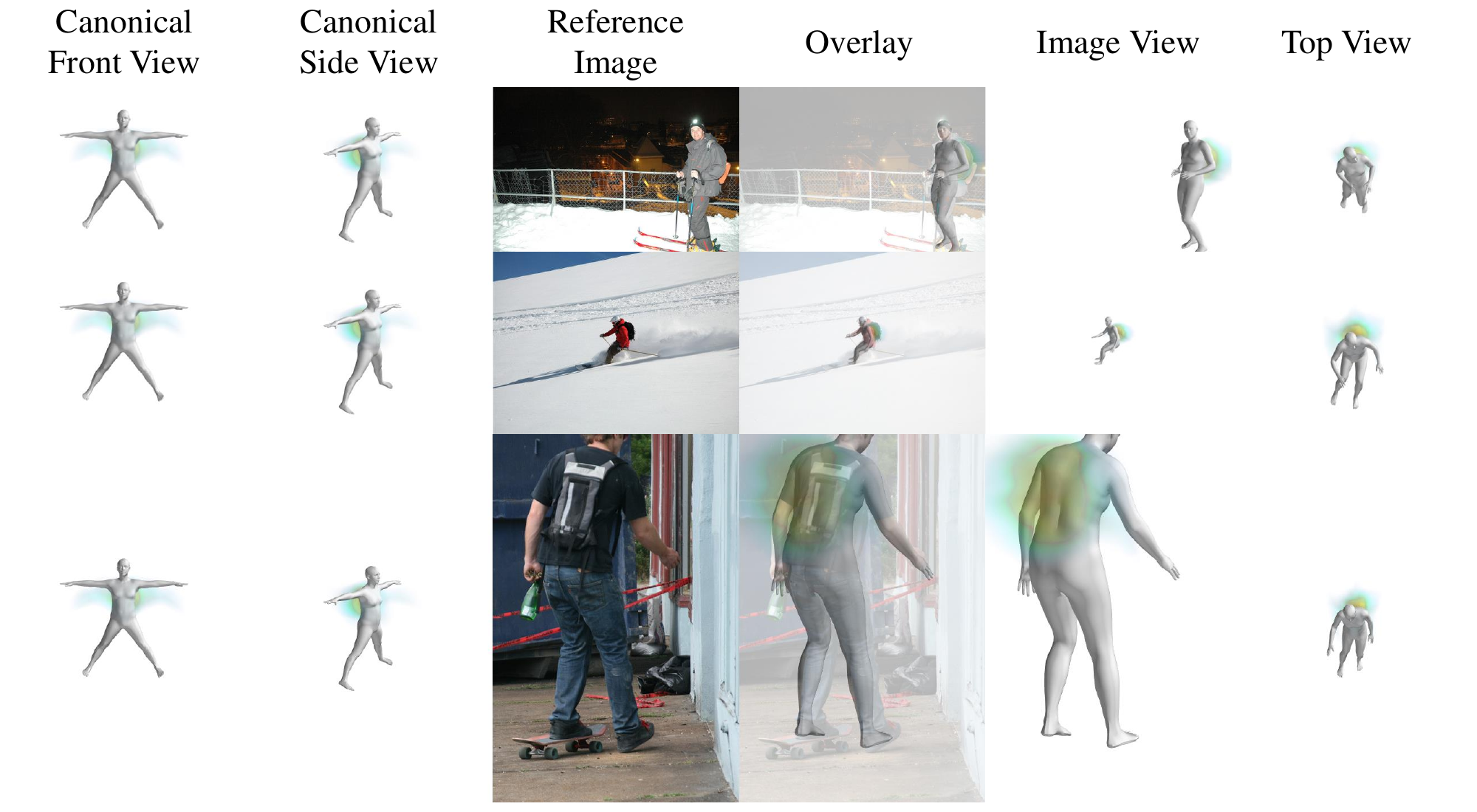}
    \caption{Qualitative results for category \textbf{backpack}.}
    \label{fig:qual backpack}
\end{figure*}

\begin{figure*}[p]
    \centering
    \includegraphics[width=2.0\columnwidth]{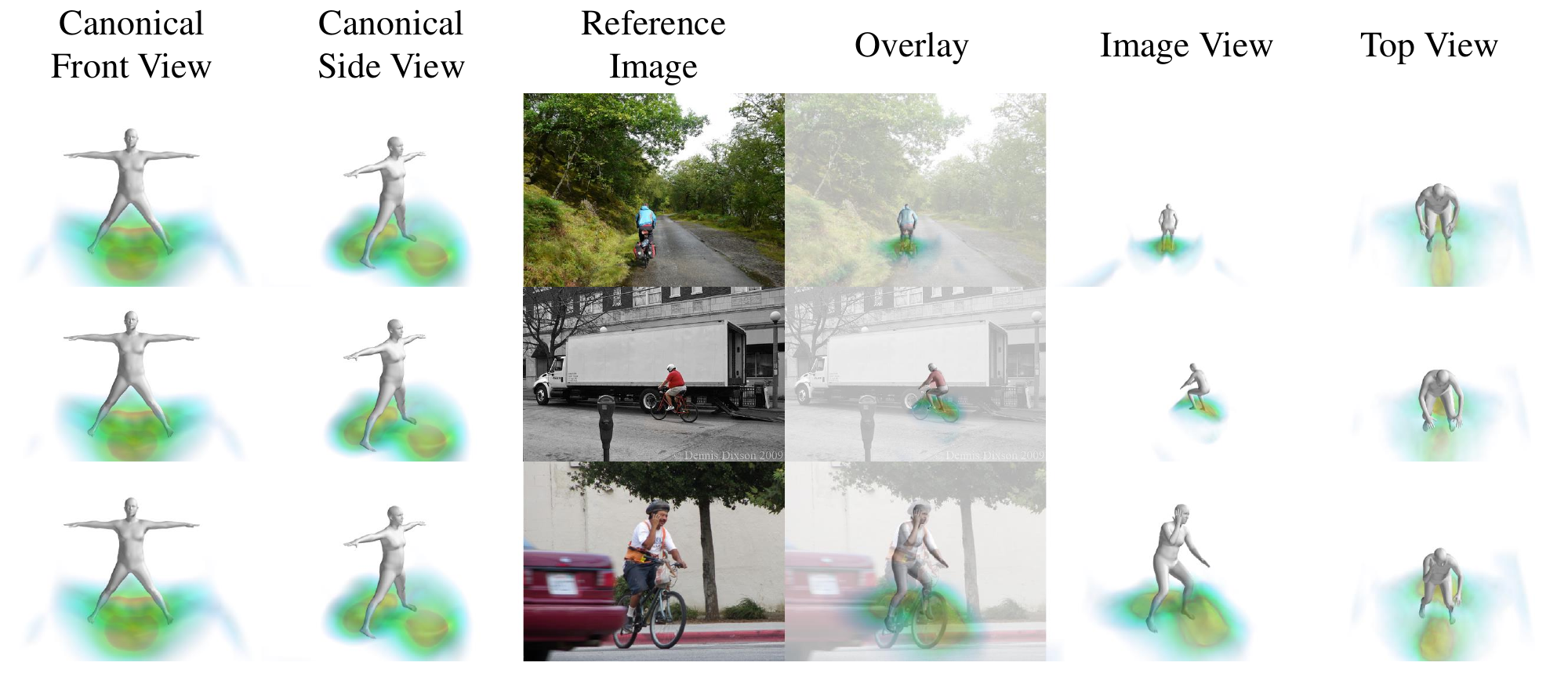}
    \caption{Qualitative results for category \textbf{bicycle}.}
    \label{fig:qual bicycle}
\end{figure*}

\begin{figure*}[p]
    \centering
    \includegraphics[width=2.0\columnwidth]{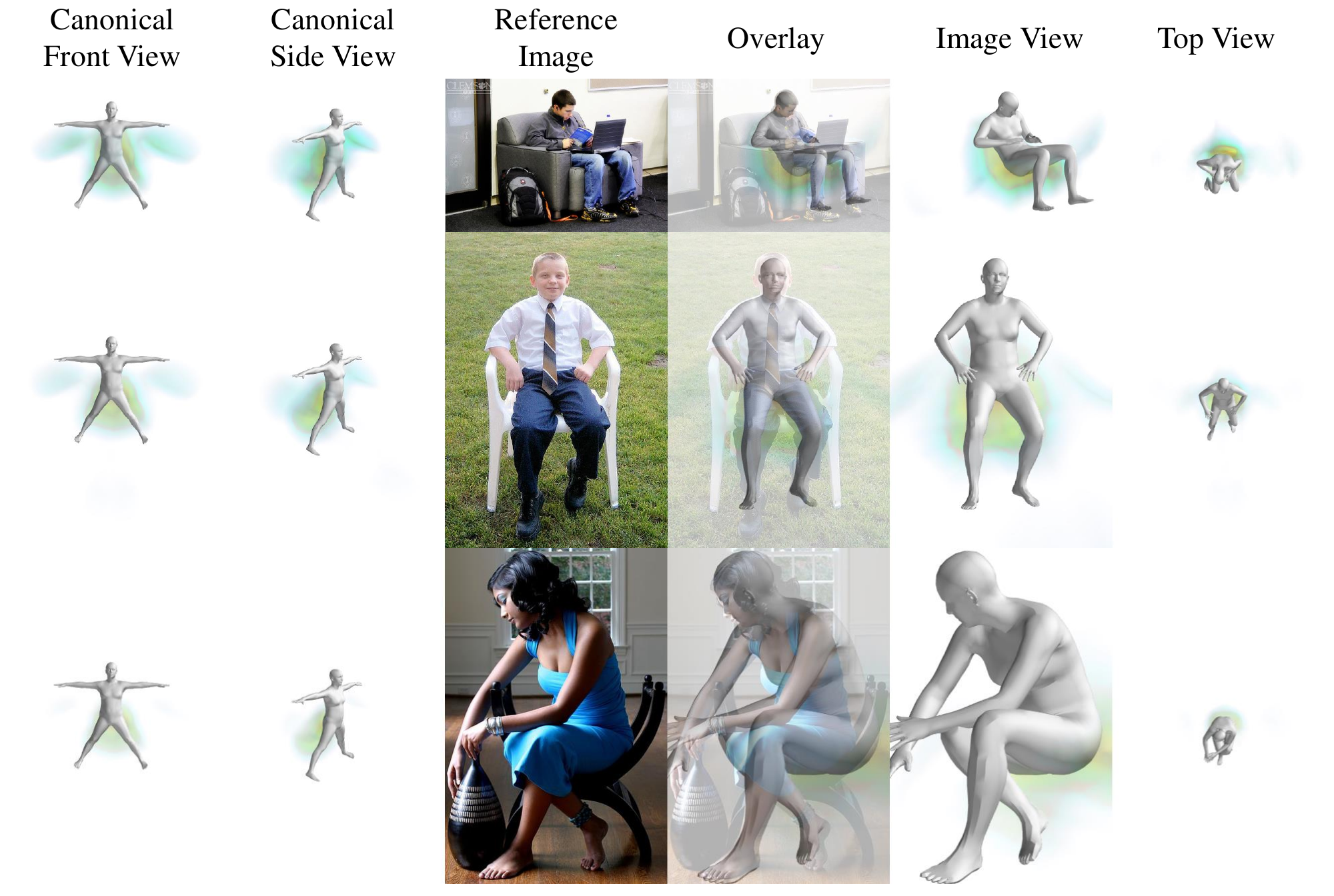}
    \caption{Qualitative results for category \textbf{chair}.}
    \label{fig:qual chair}
\end{figure*}

\begin{figure*}[p]
    \centering
    \includegraphics[width=2.0\columnwidth]{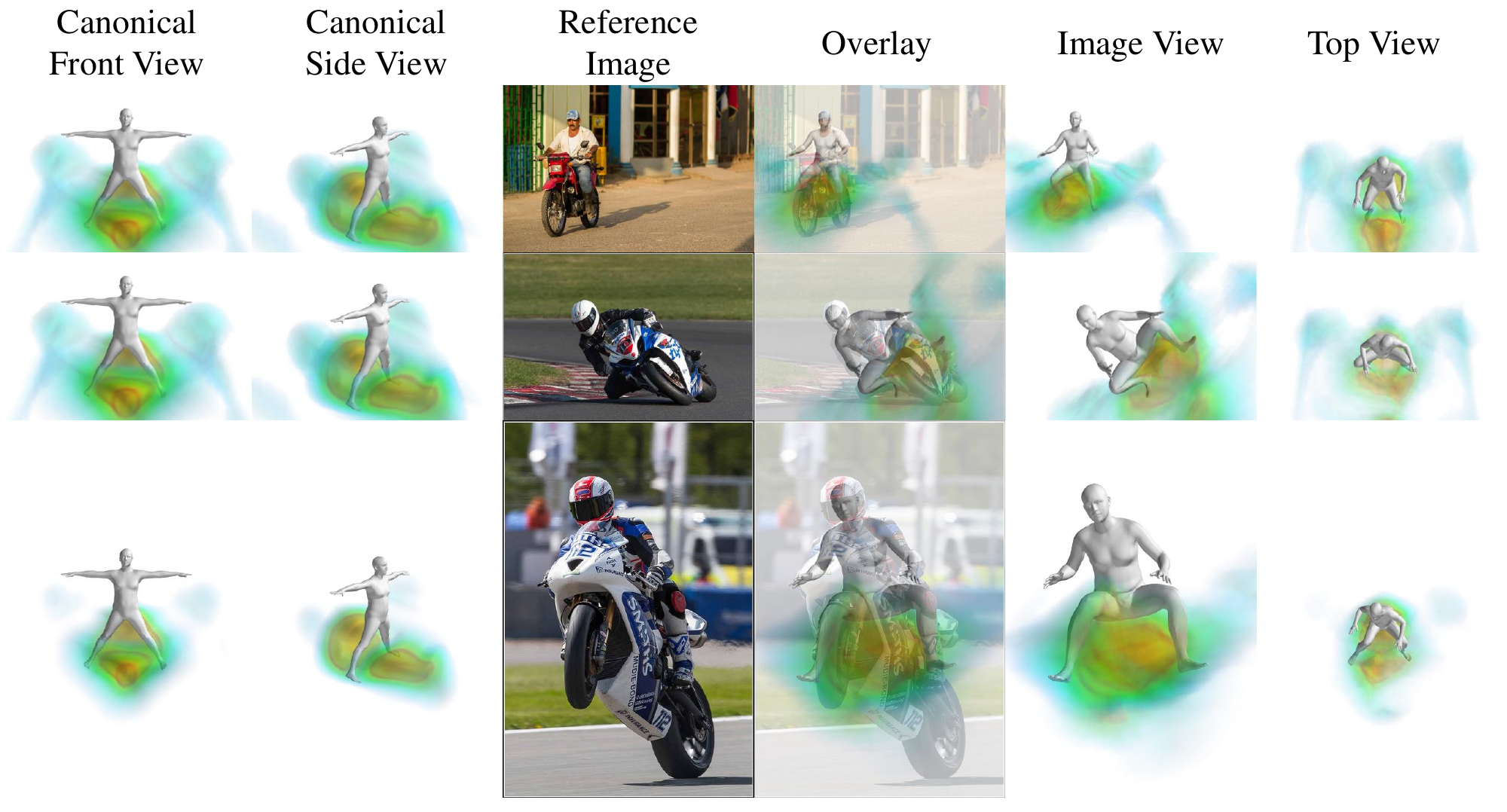}
    \caption{Qualitative results for category \textbf{motorcycle}.}
    \label{fig:qual motorcycle}
\end{figure*}

\begin{figure*}[p]
    \centering
    \includegraphics[width=2.0\columnwidth]{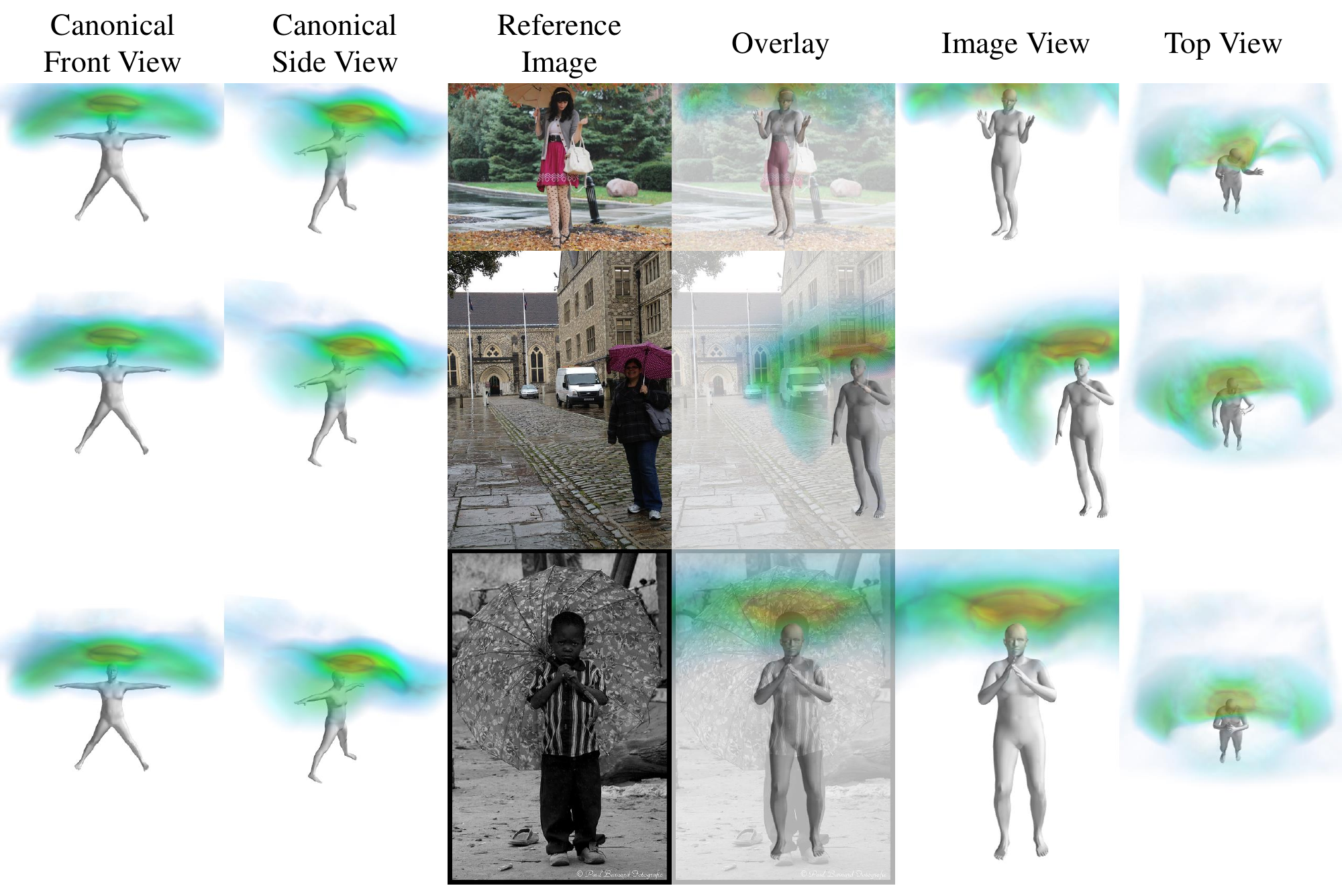}
    \caption{Qualitative results for category \textbf{umbrella}.}
    \label{fig:qual umbrella}
\end{figure*}

\begin{figure*}[p]
    \centering
    \includegraphics[width=2.0\columnwidth]{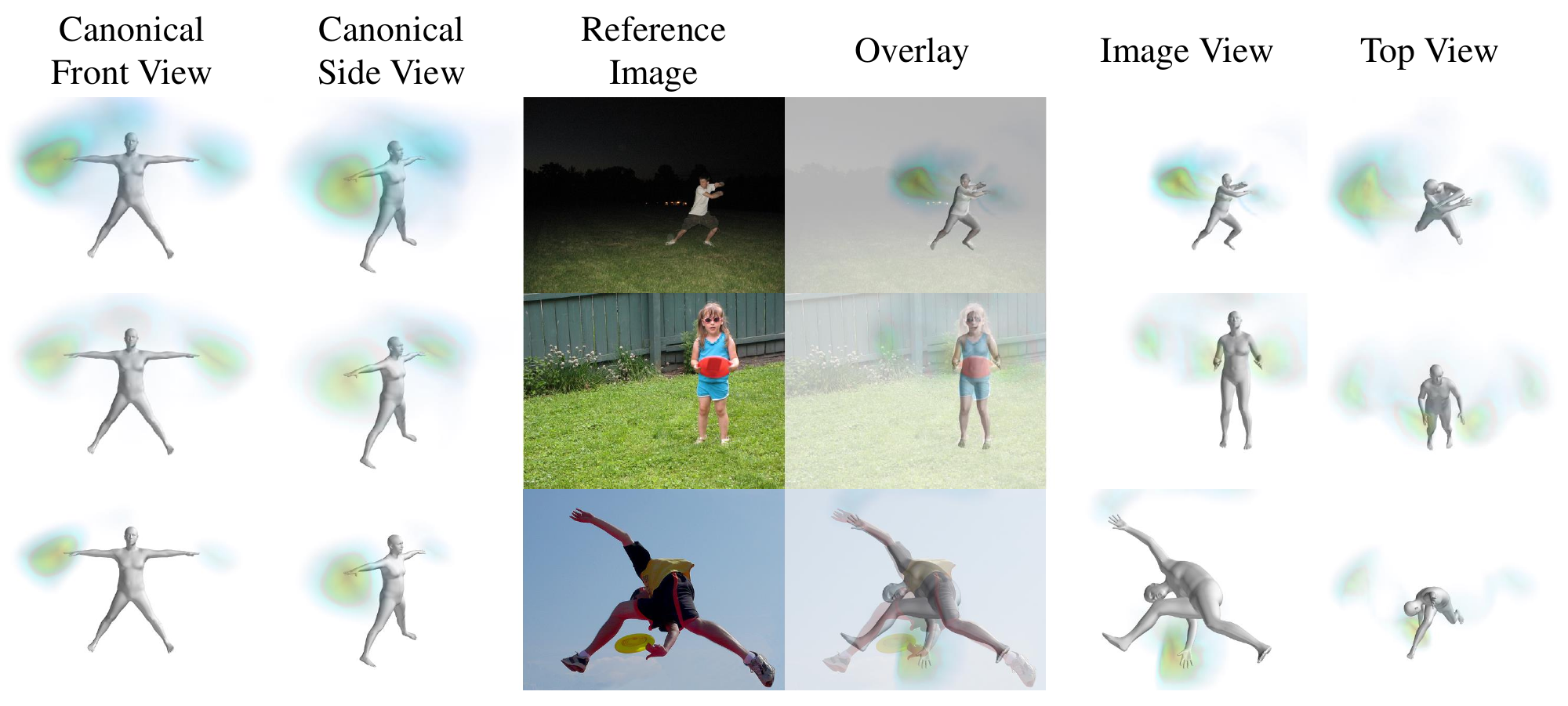}
    \caption{Qualitative results for category \textbf{frisbee}.}
    \label{fig:qual frisbee}
\end{figure*}

\begin{figure*}[p]
    \centering
    \includegraphics[width=2.0\columnwidth]{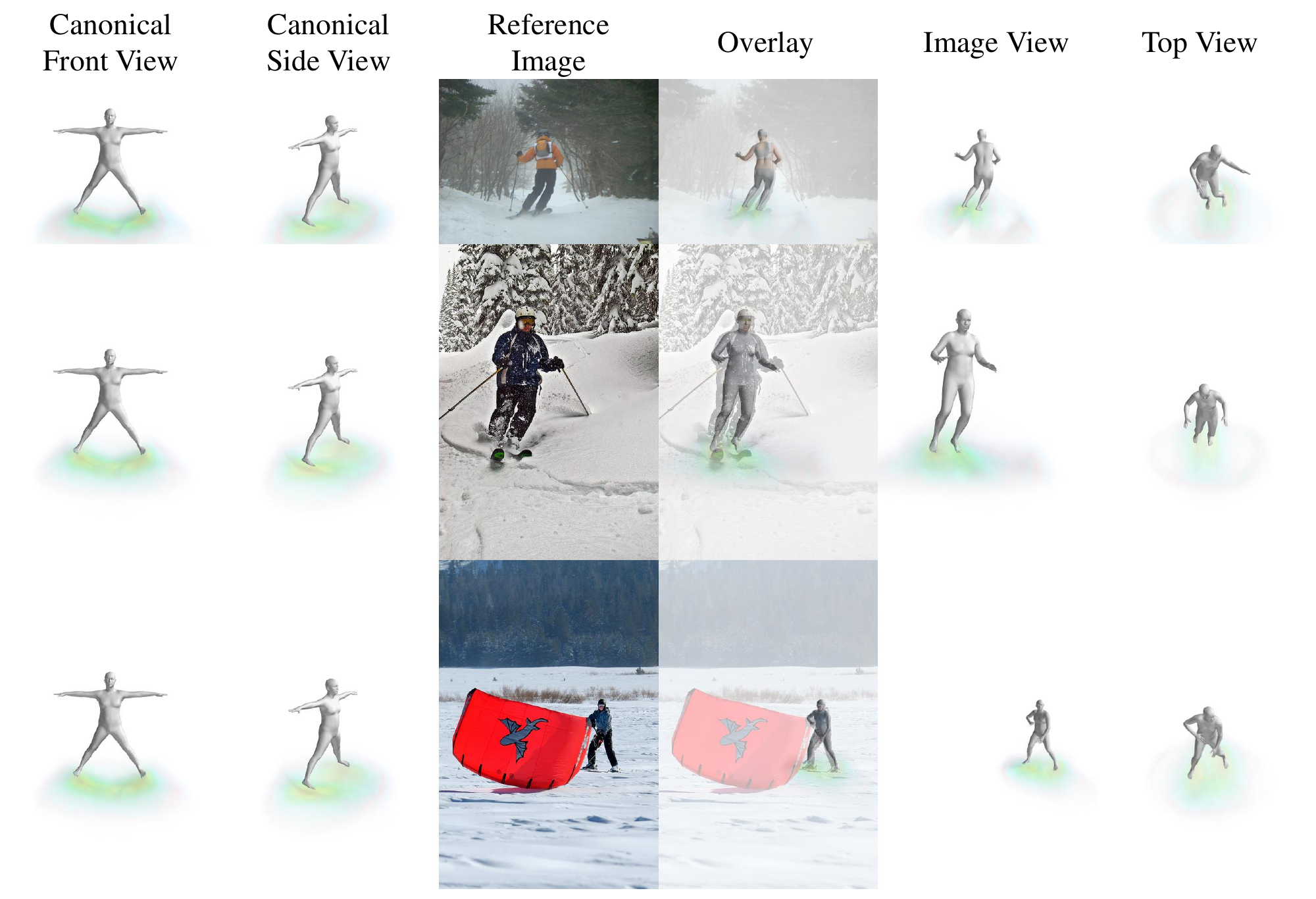}
    \caption{Qualitative results for category \textbf{skis}.}
    \label{fig:qual skis}
\end{figure*}

\begin{figure*}[p]
    \centering
    \includegraphics[width=2.0\columnwidth]{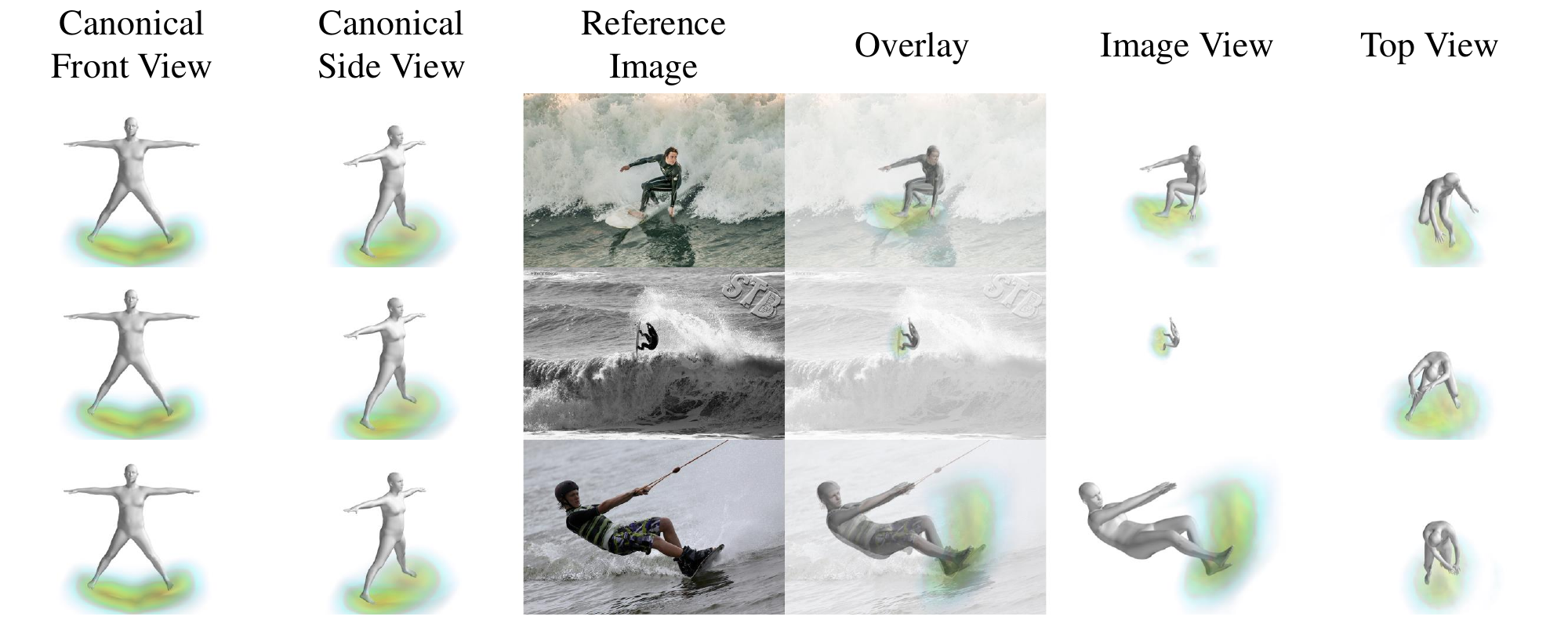}
    \caption{Qualitative results for category \textbf{surfboard}.}
    \label{fig:qual surfboard}
\end{figure*}

\begin{figure*}[p]
    \centering
    \includegraphics[width=2.0\columnwidth]{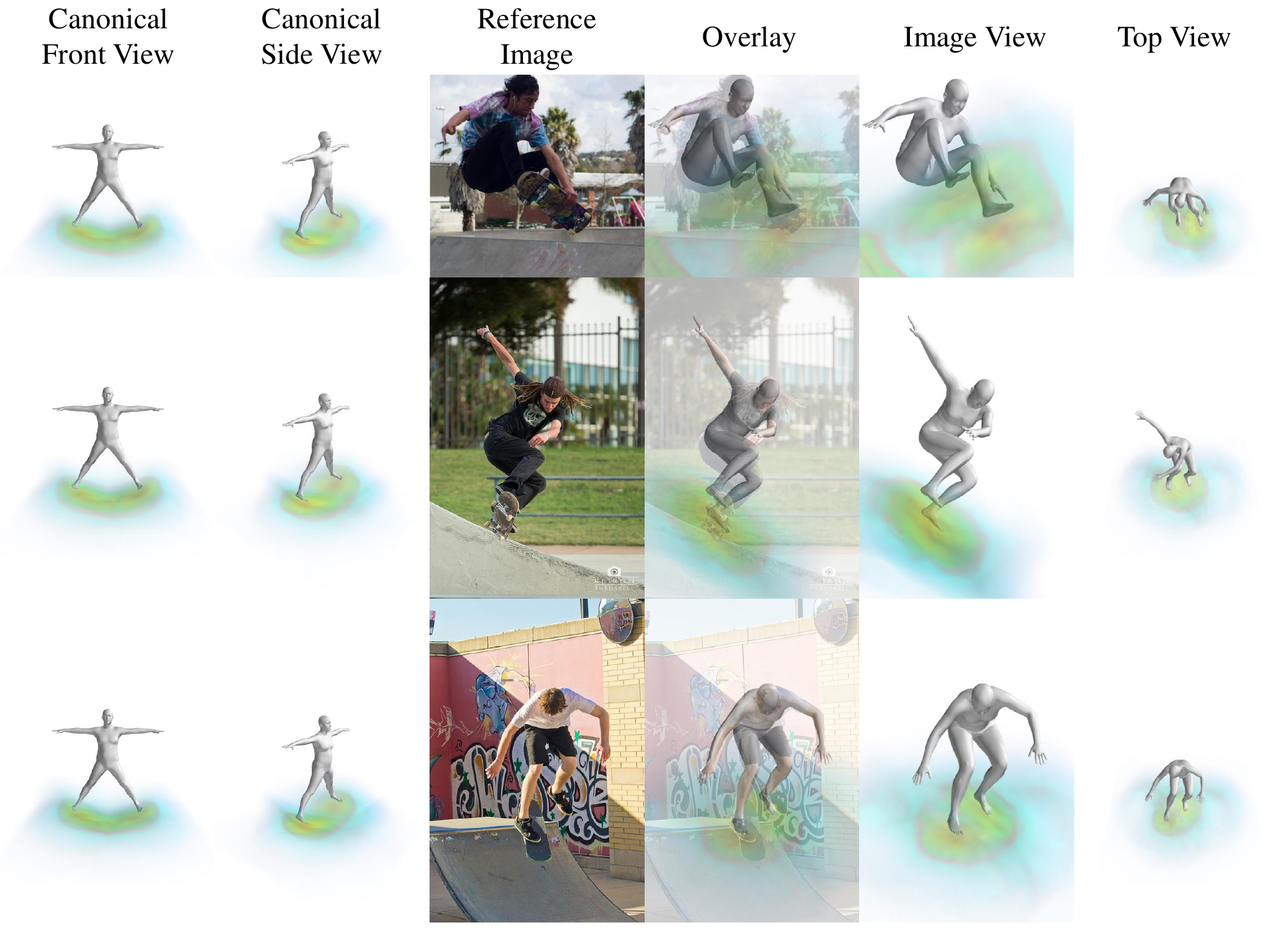}
    \caption{Qualitative results for category \textbf{skateboard}.}
    \label{fig:qual skateboard}
\end{figure*}

\begin{figure*}[p]
    \centering
    \includegraphics[width=2.0\columnwidth]{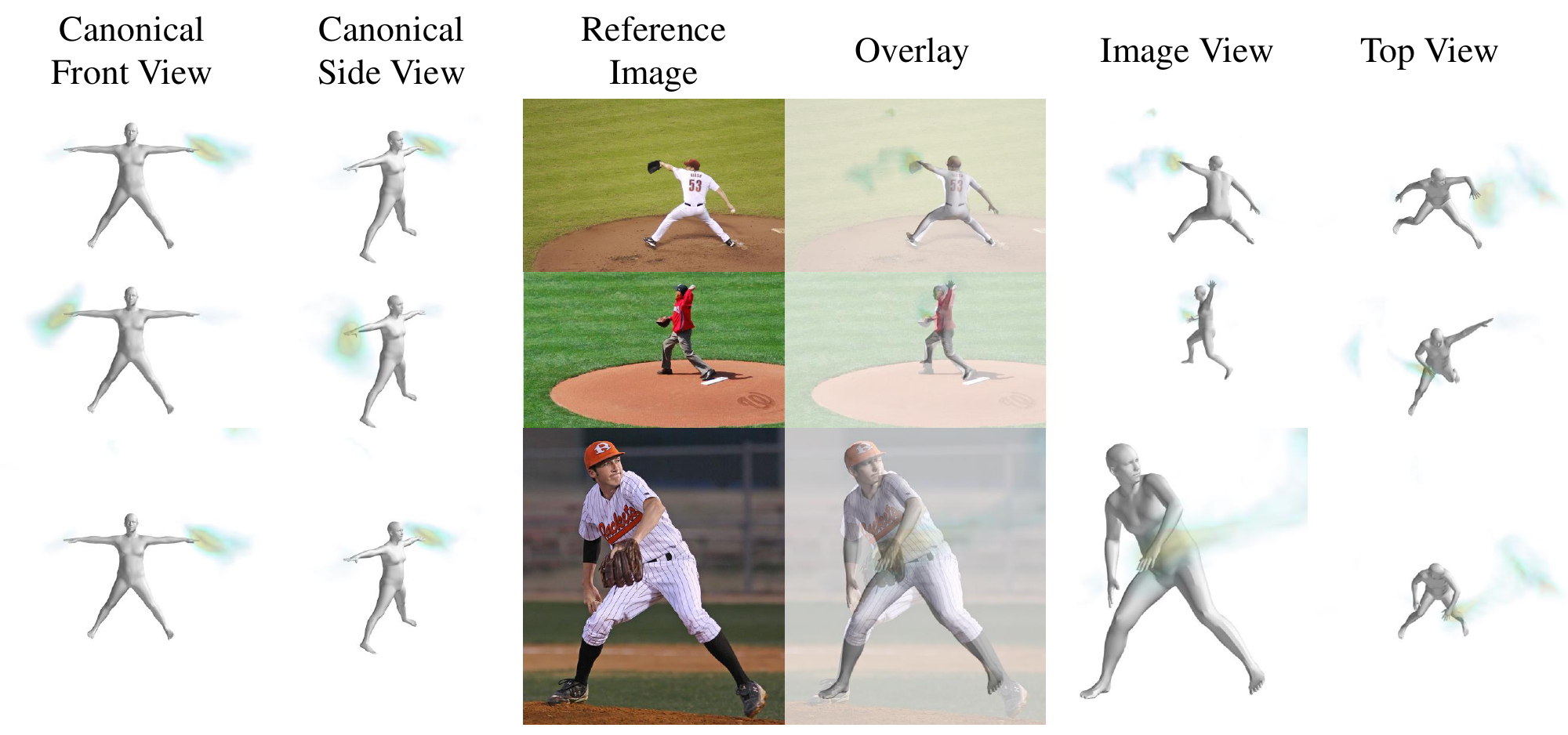}
    \caption{Qualitative results for category \textbf{baseball glove}.}
    \label{fig:qual baseball glove}
\end{figure*}

\begin{figure*}[p]
    \centering
    \includegraphics[width=2.0\columnwidth]{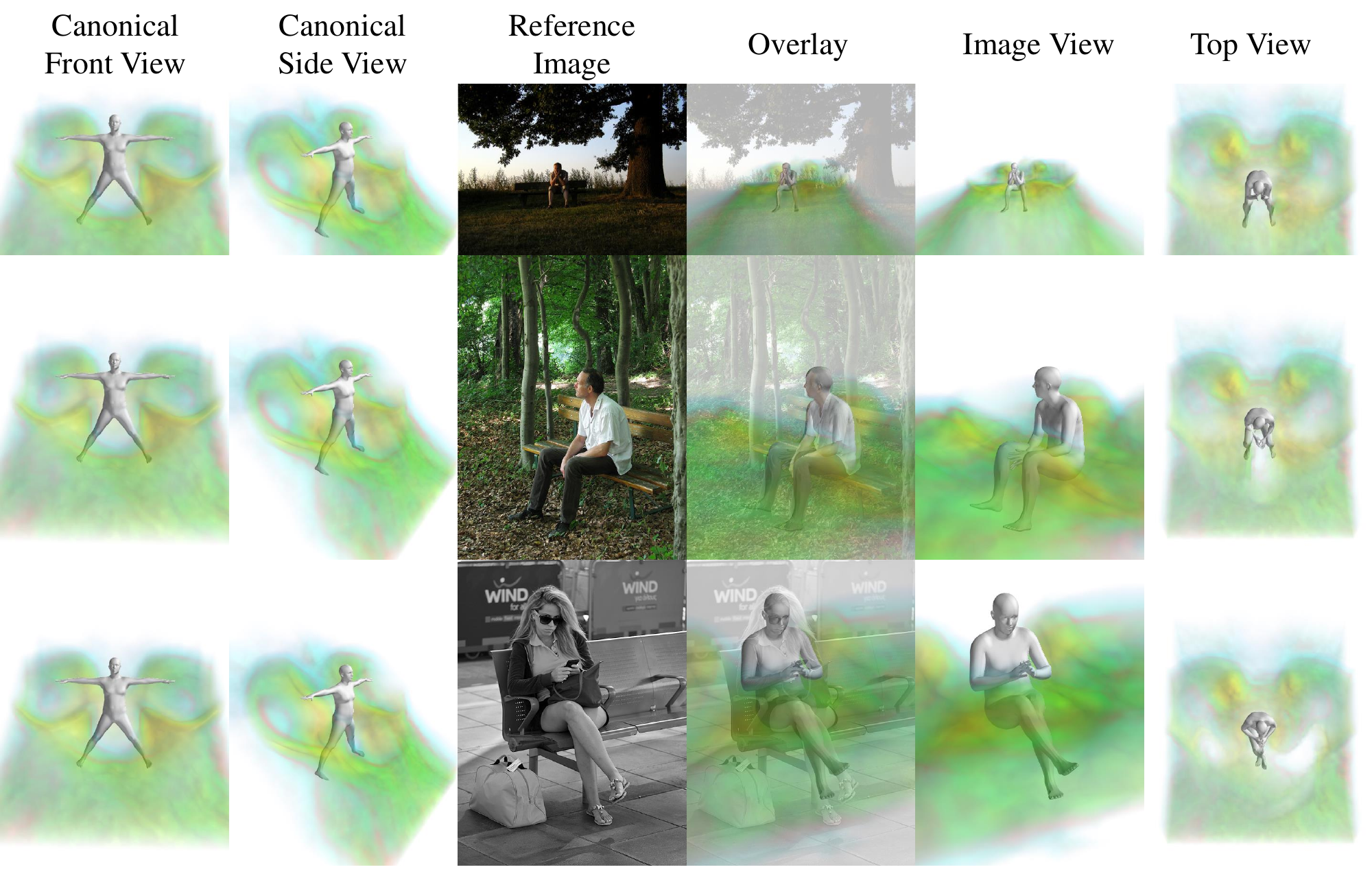}
    \caption{Qualitative results for category \textbf{bench}.}
    \label{fig:qual bench}
\end{figure*}

\begin{figure*}[p]
    \centering
    \includegraphics[width=2.0\columnwidth]{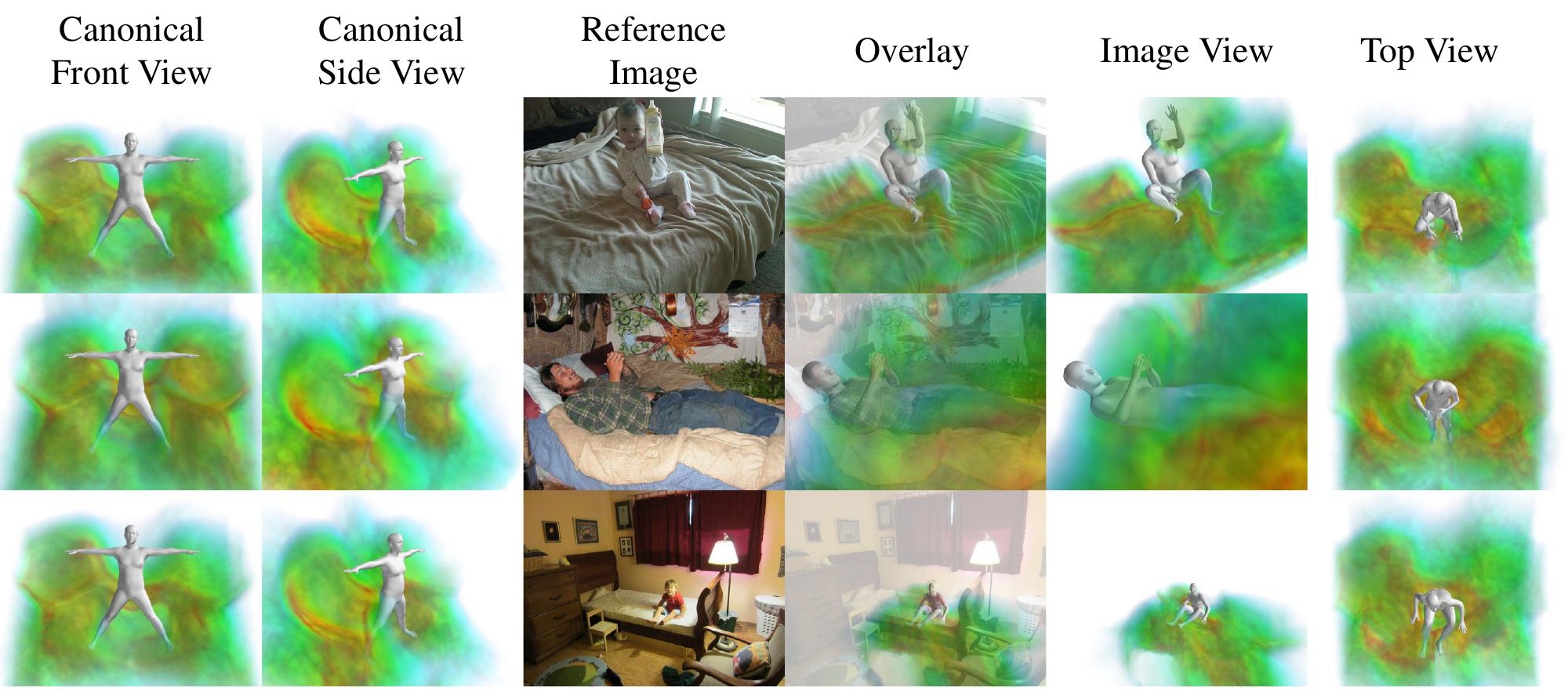}
    \caption{Qualitative results for category \textbf{bed}.}
    \label{fig:qual bed}
\end{figure*}

\begin{figure*}[p]
    \centering
    \includegraphics[width=2.0\columnwidth]{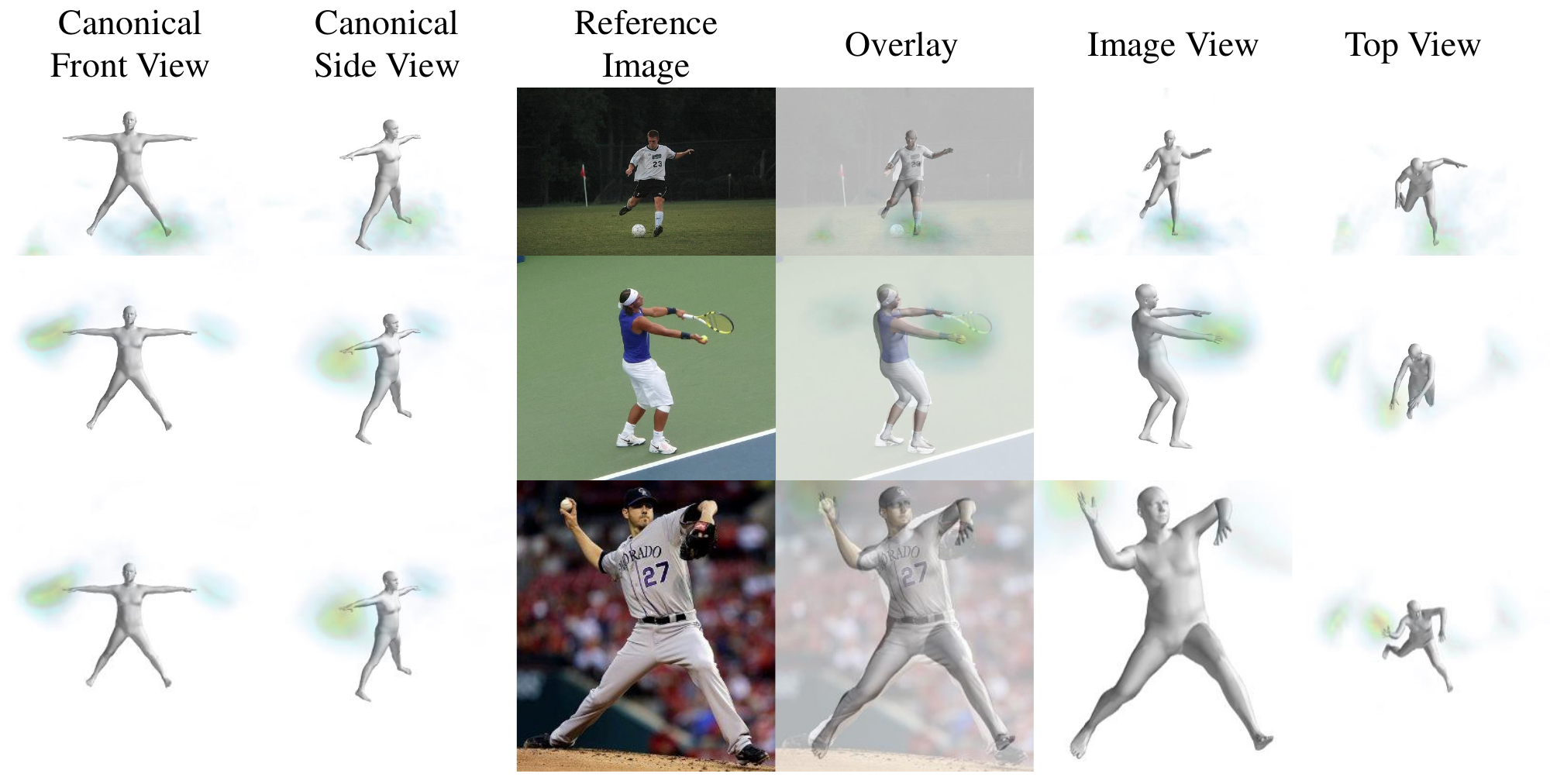}
    \caption{Qualitative results for category \textbf{sports ball}.}
    \label{fig:qual sports ball}
\end{figure*}

\begin{figure*}[p]
    \centering
    \includegraphics[width=2.0\columnwidth]{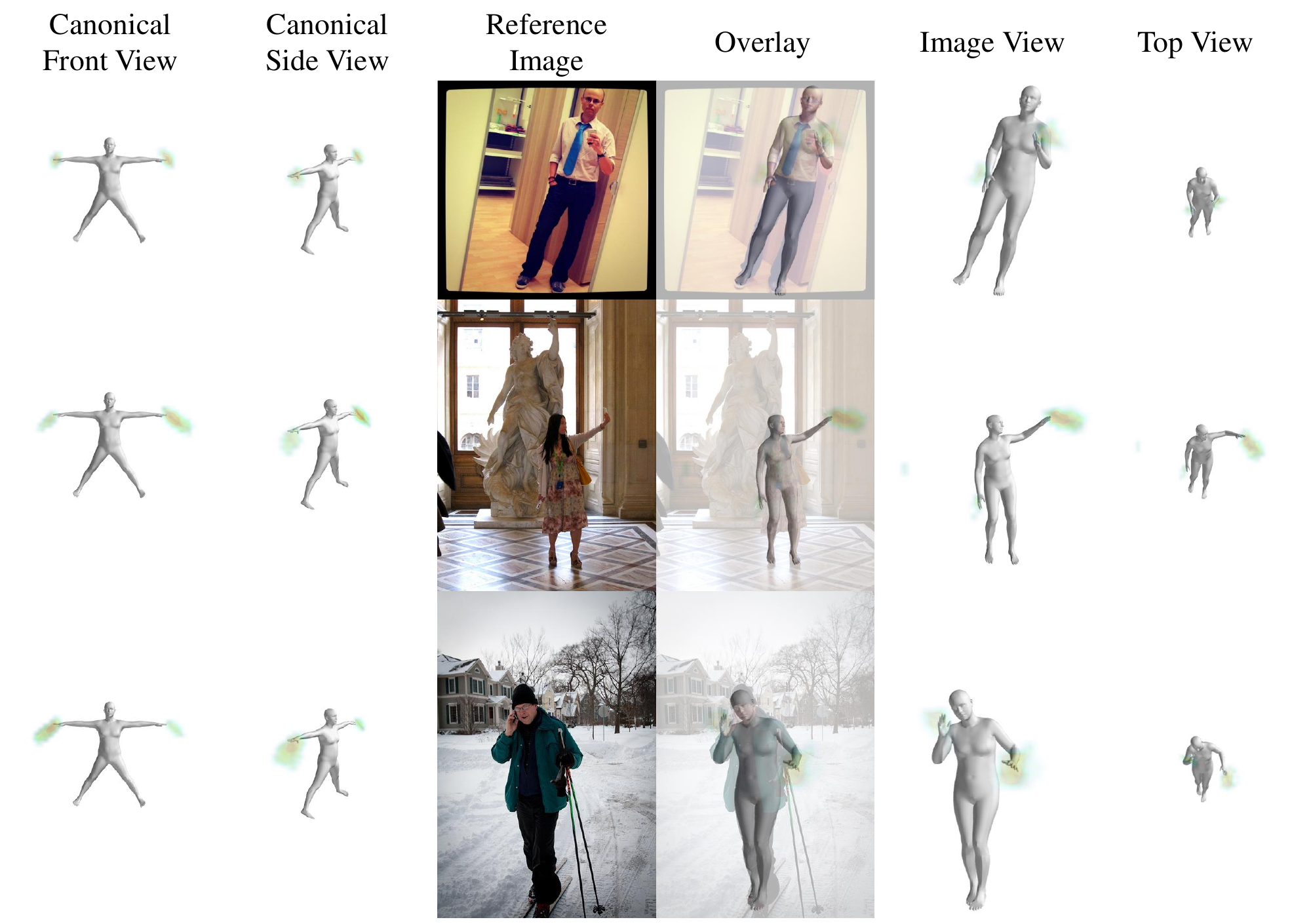}
    \caption{Qualitative results for category \textbf{cell phone}.}
    \label{fig:qual cell phone}
\end{figure*}

\begin{figure*}[p]
    \centering
    \includegraphics[width=2.0\columnwidth]{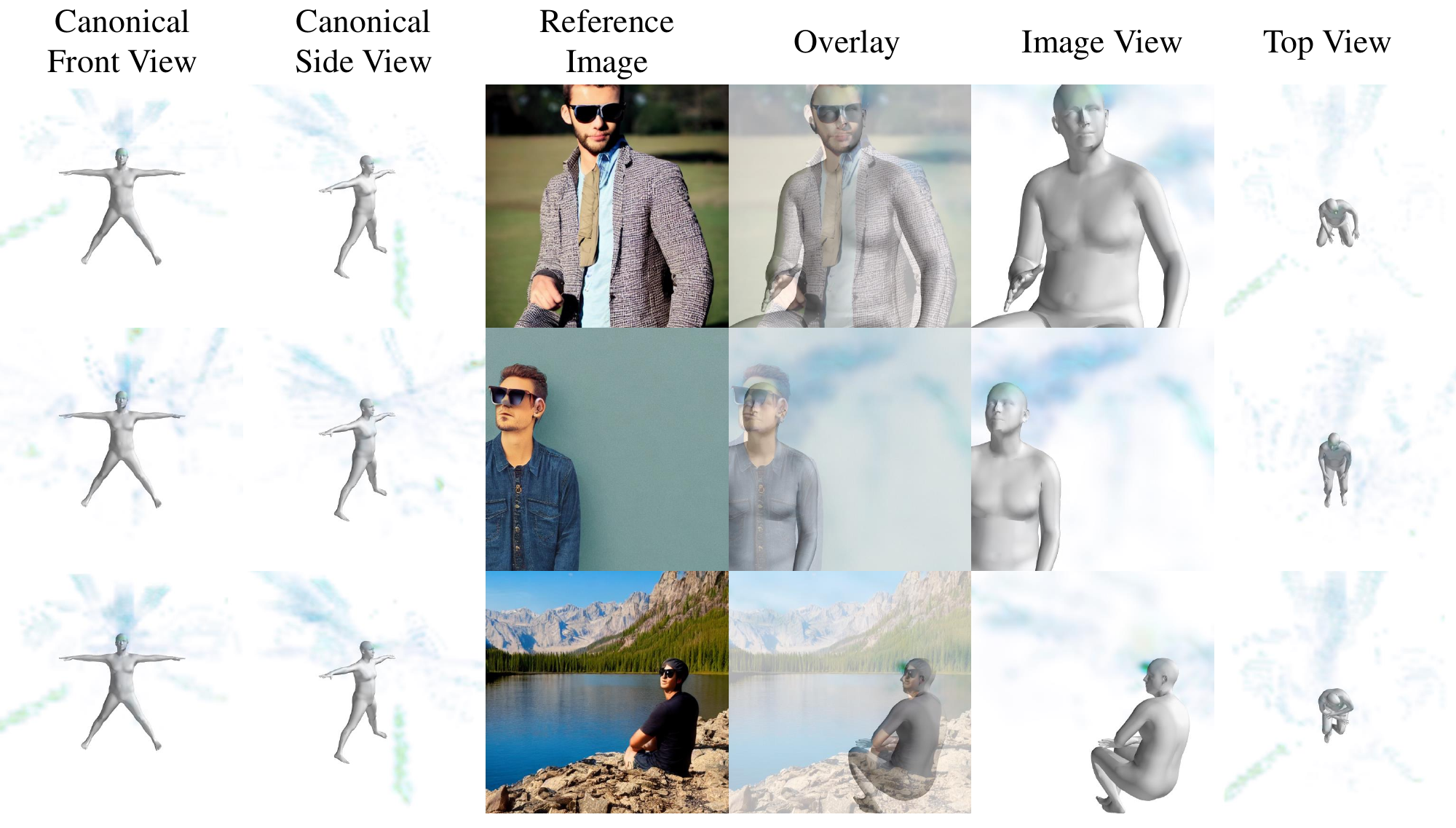}
    \caption{Qualitative results for category \textbf{sunglasses}.}
    \label{fig:qual sunglasses}
\end{figure*}

\begin{figure*}[p]
    \centering
    \includegraphics[width=2.0\columnwidth]{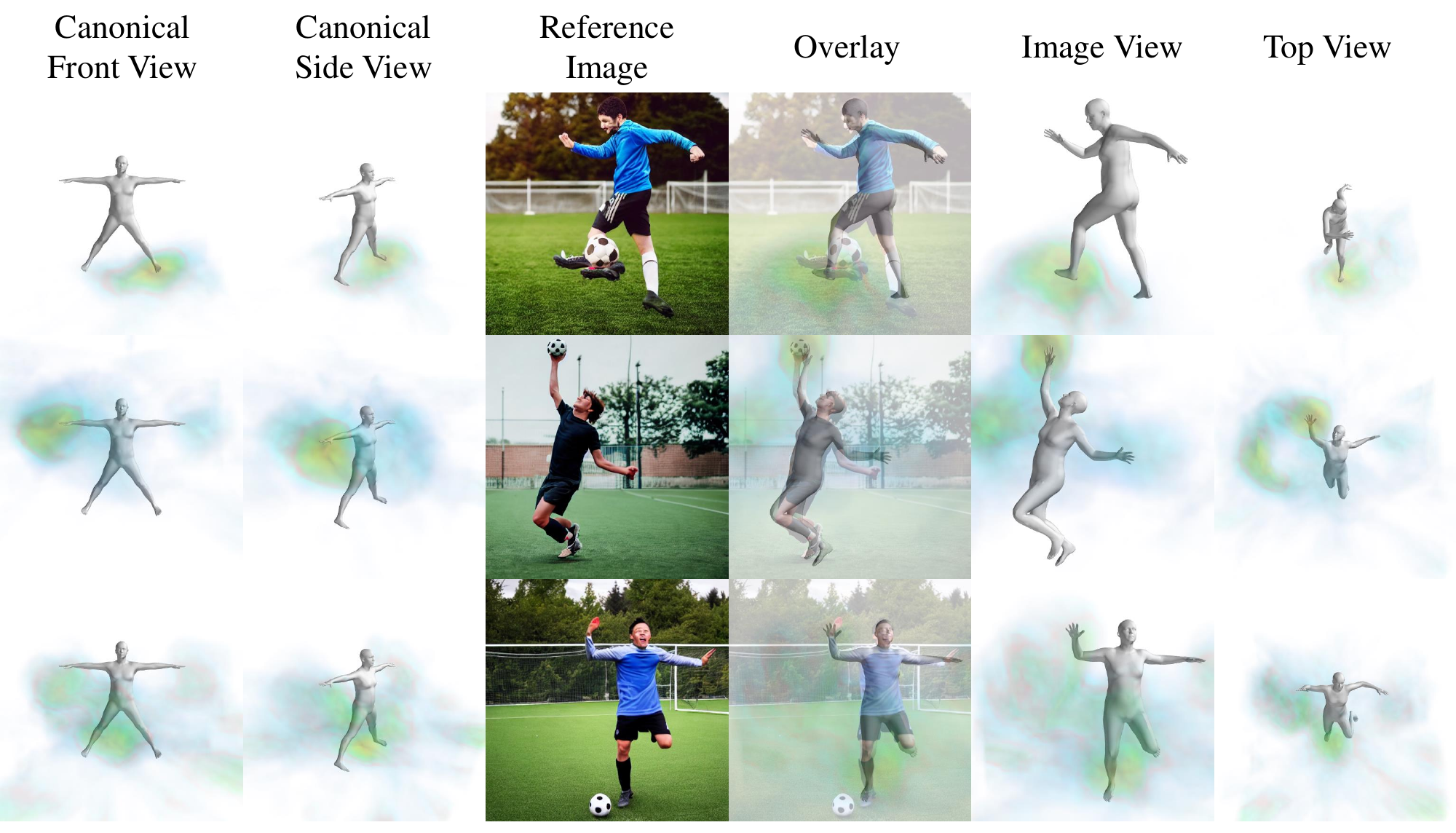}
    \caption{Qualitative results for category \textbf{soccer ball}.}
    \label{fig:qual soccer ball}
\end{figure*}

\begin{figure*}[p]
    \centering
    \includegraphics[width=2.0\columnwidth]{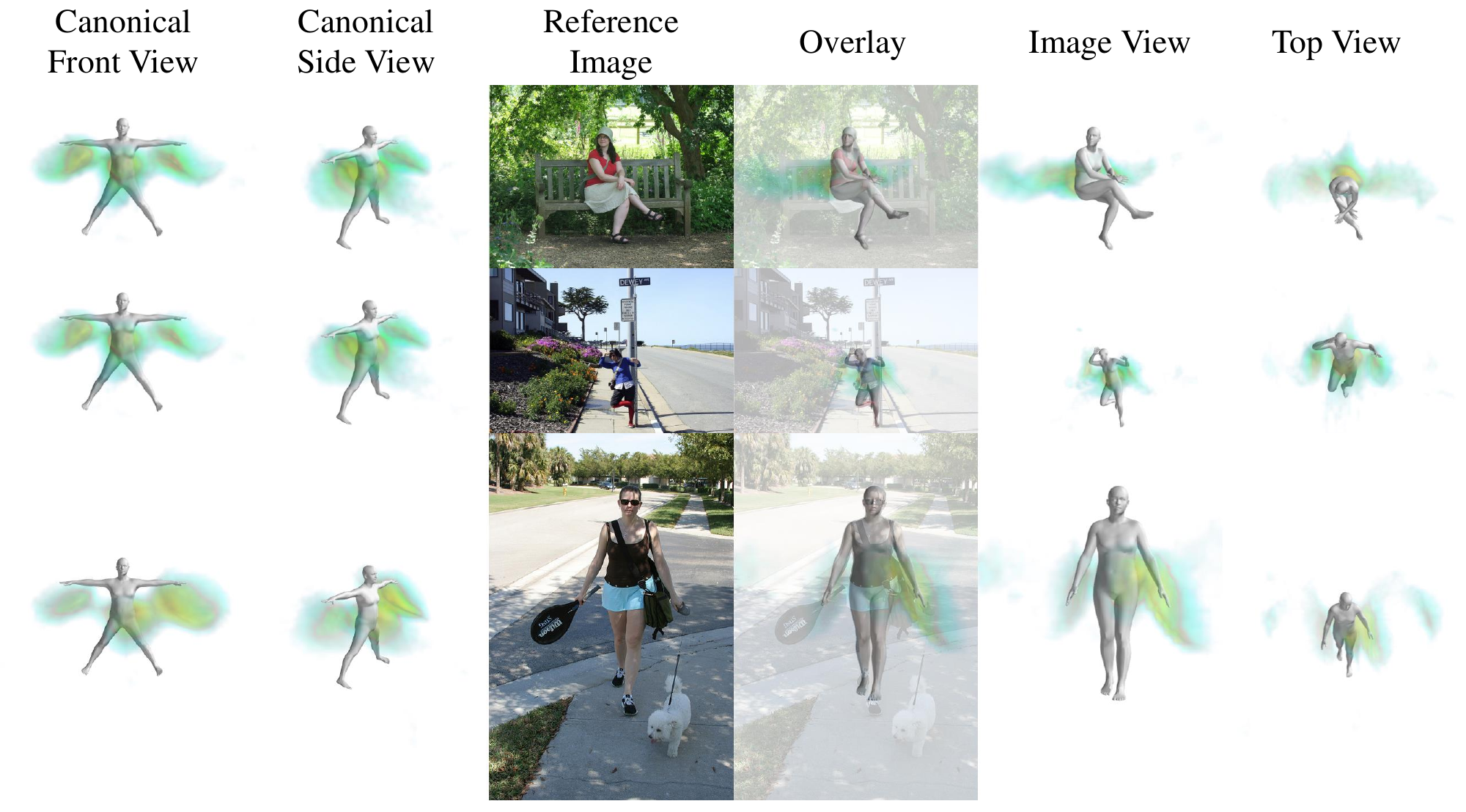}
    \caption{Qualitative results for category \textbf{handbag}.}
    \label{fig:qual handbag}
\end{figure*}

\begin{figure*}[p]
    \centering
    \includegraphics[width=2.0\columnwidth]{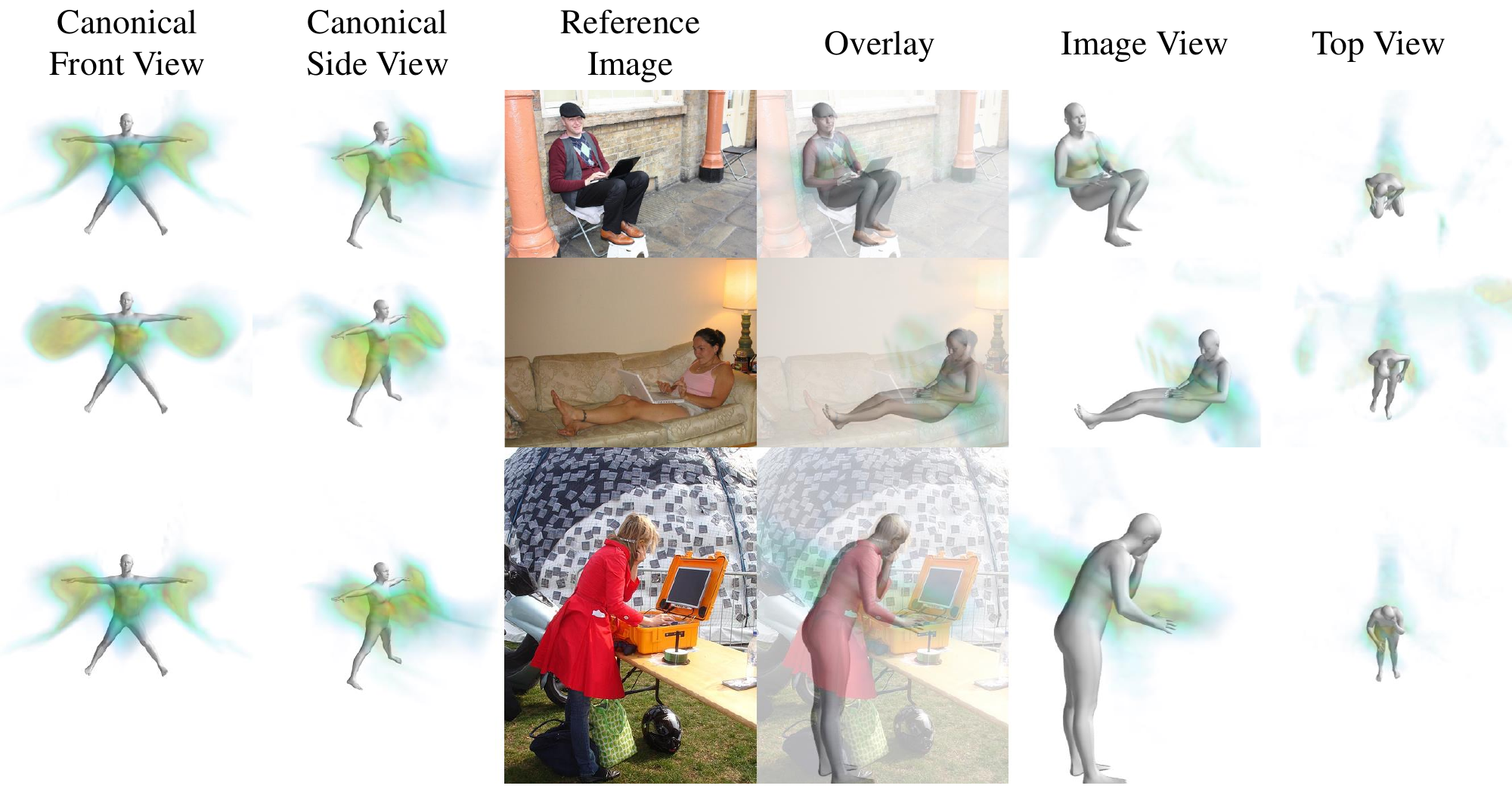}
    \caption{Qualitative results for category \textbf{laptop}.}
    \label{fig:qual laptop}
\end{figure*}

\begin{figure*}[p]
    \centering
    \includegraphics[width=2.0\columnwidth]{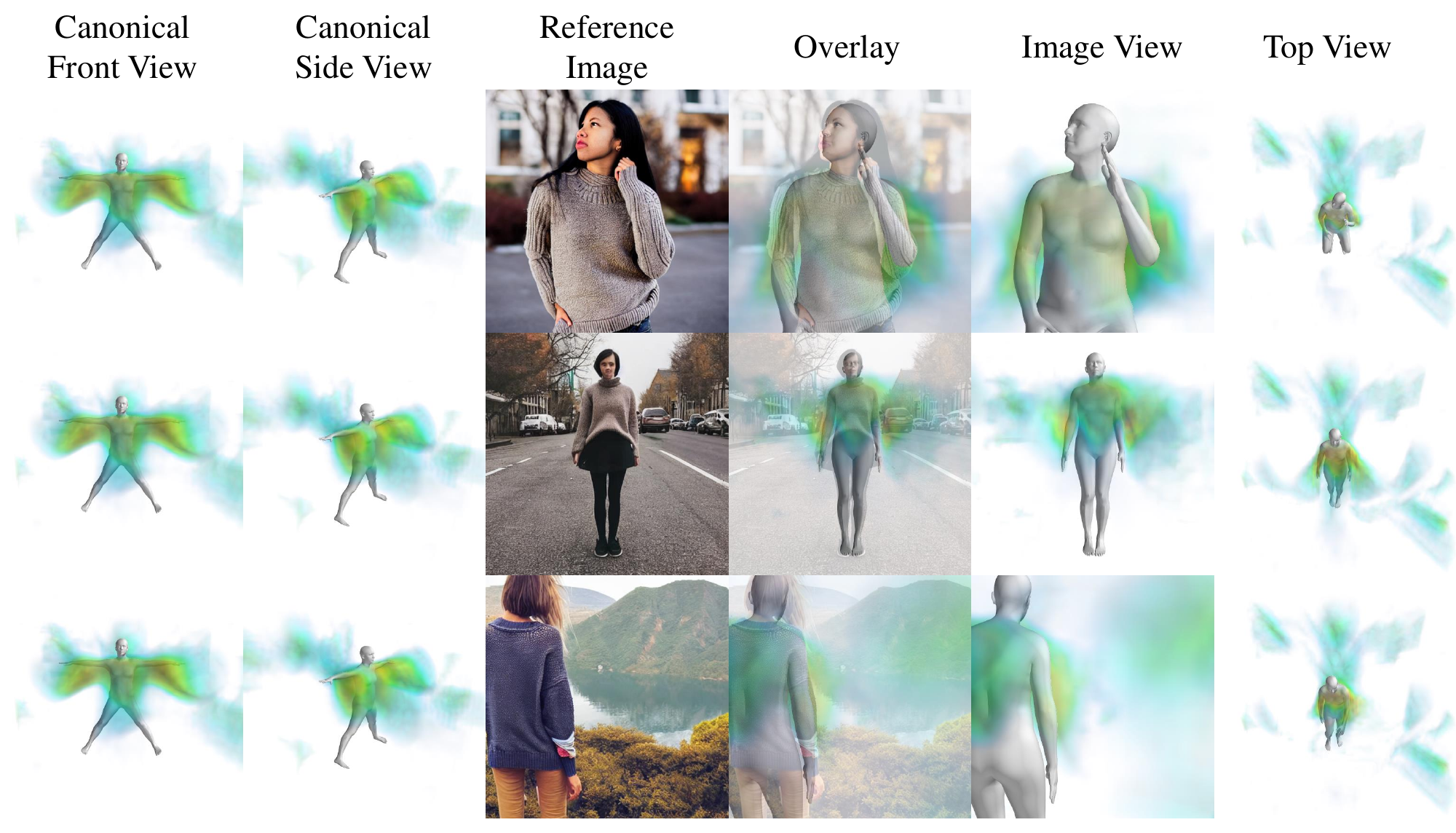}
    \caption{Qualitative results for category \textbf{sweater}.}
    \label{fig:qual sweater}
\end{figure*}

\begin{figure*}[p]
    \centering
    \includegraphics[width=2.0\columnwidth]{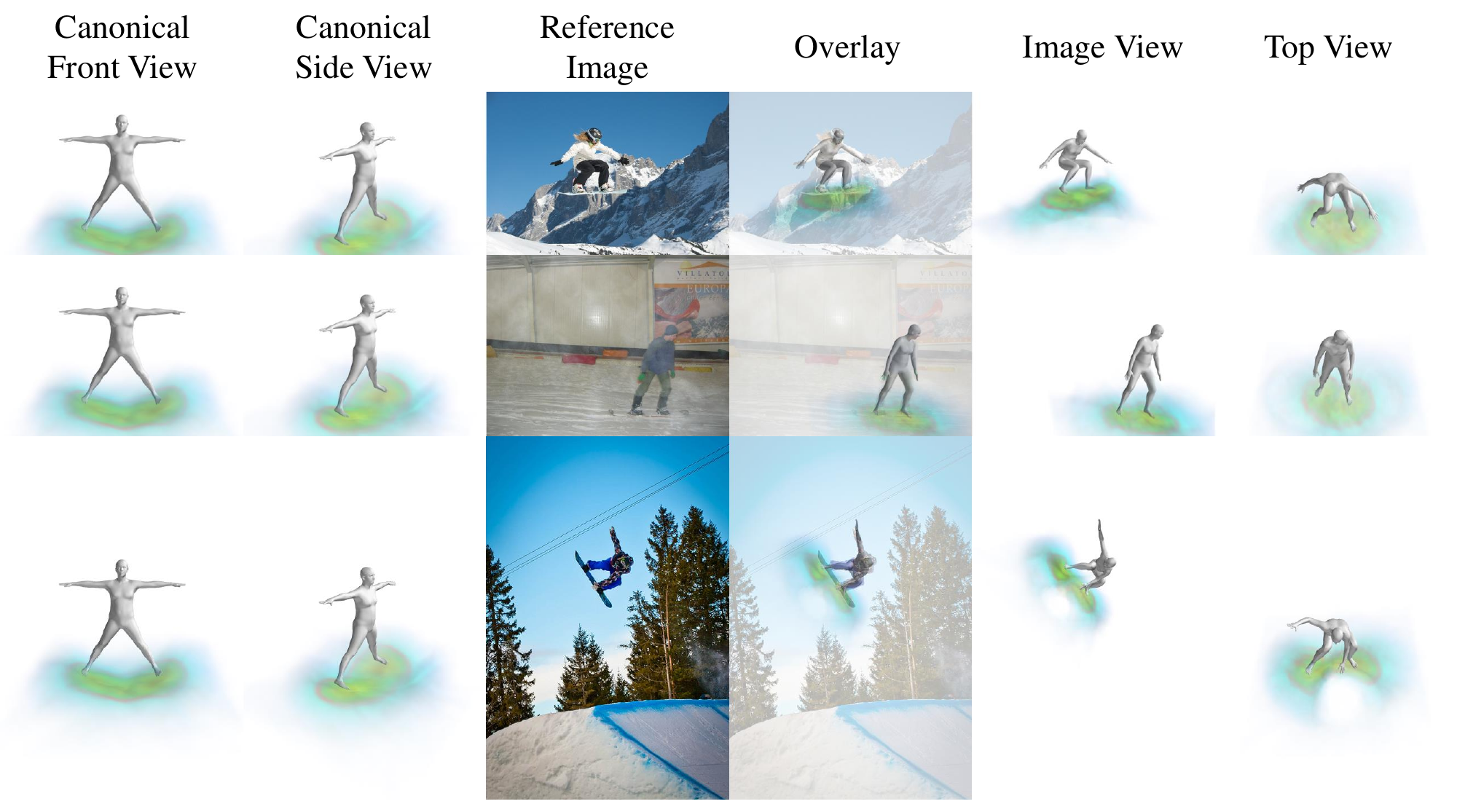}
    \caption{Qualitative results for category \textbf{snowboard}.}
    \label{fig:qual snowboard}
\end{figure*}

\begin{figure*}[p]
    \centering
    \includegraphics[width=2.0\columnwidth]{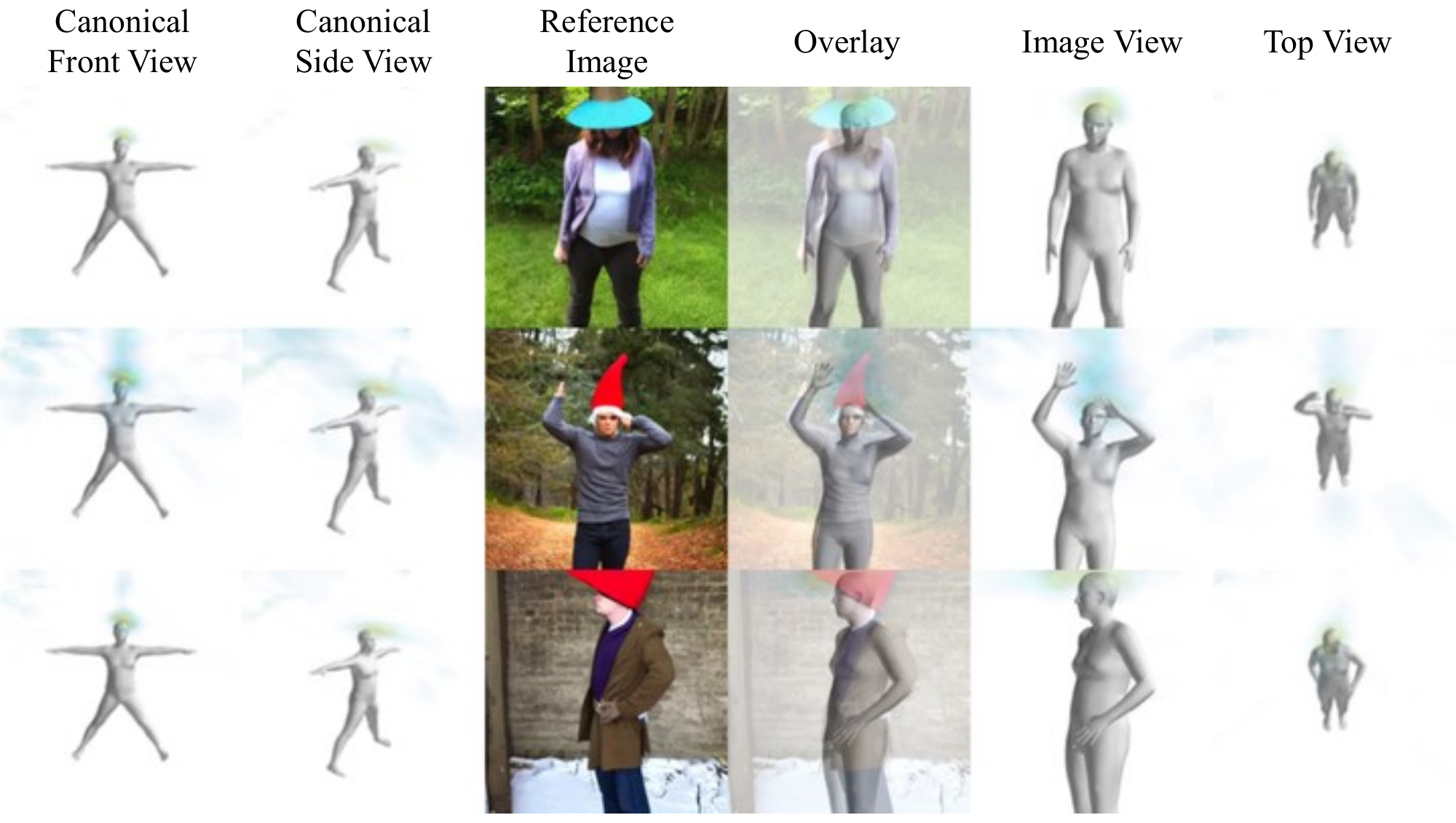}
    \caption{Qualitative results for category \textbf{hat}.}
    \label{fig:qual hat}
\end{figure*}

\begin{figure*}[p]
    \centering
    \includegraphics[width=2.0\columnwidth]{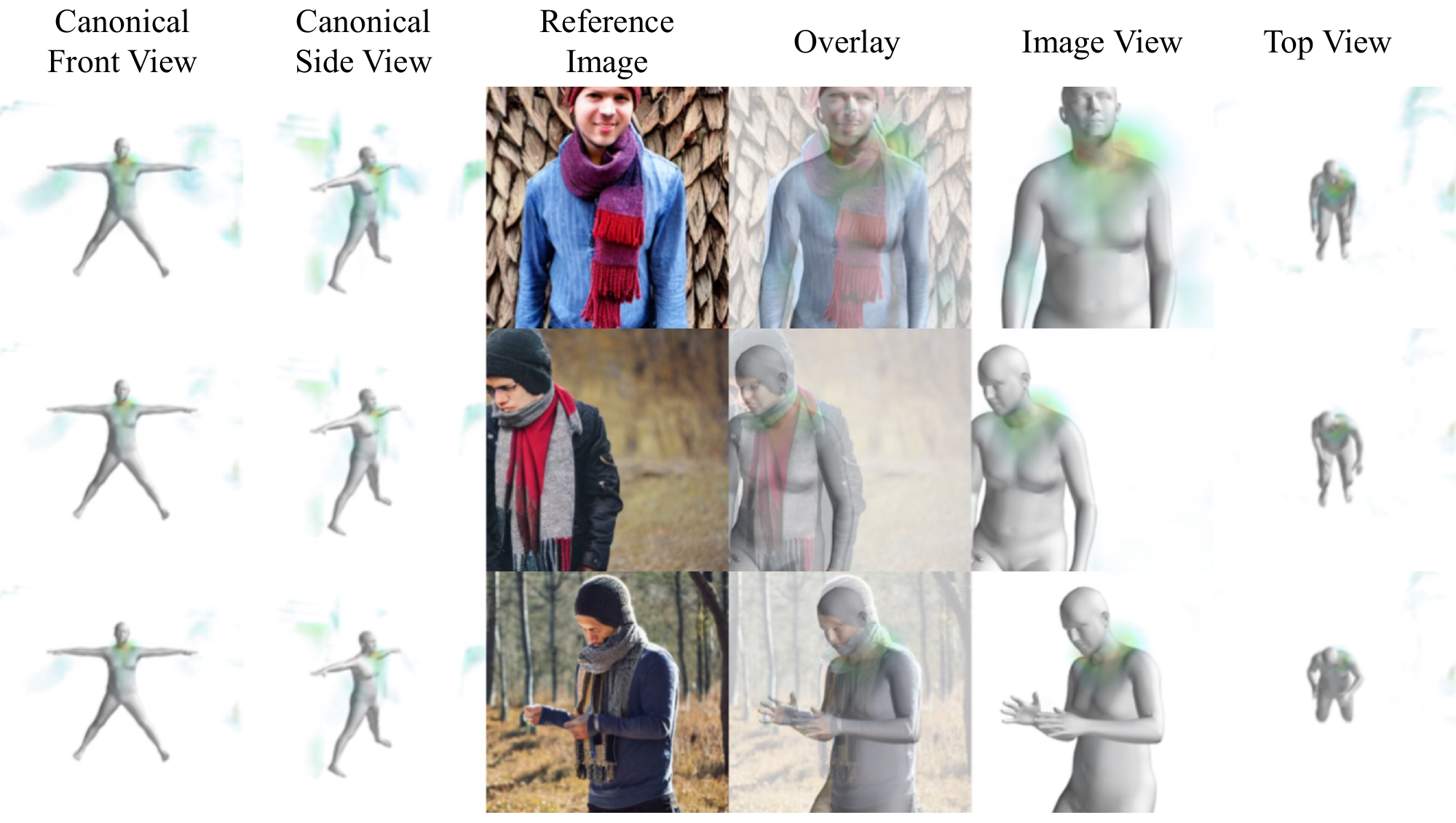}
    \caption{Qualitative results for category \textbf{scarf}.}
    \label{fig:qual scarf}
\end{figure*}

\begin{figure*}[p]
    \centering
    \includegraphics[width=2.0\columnwidth]{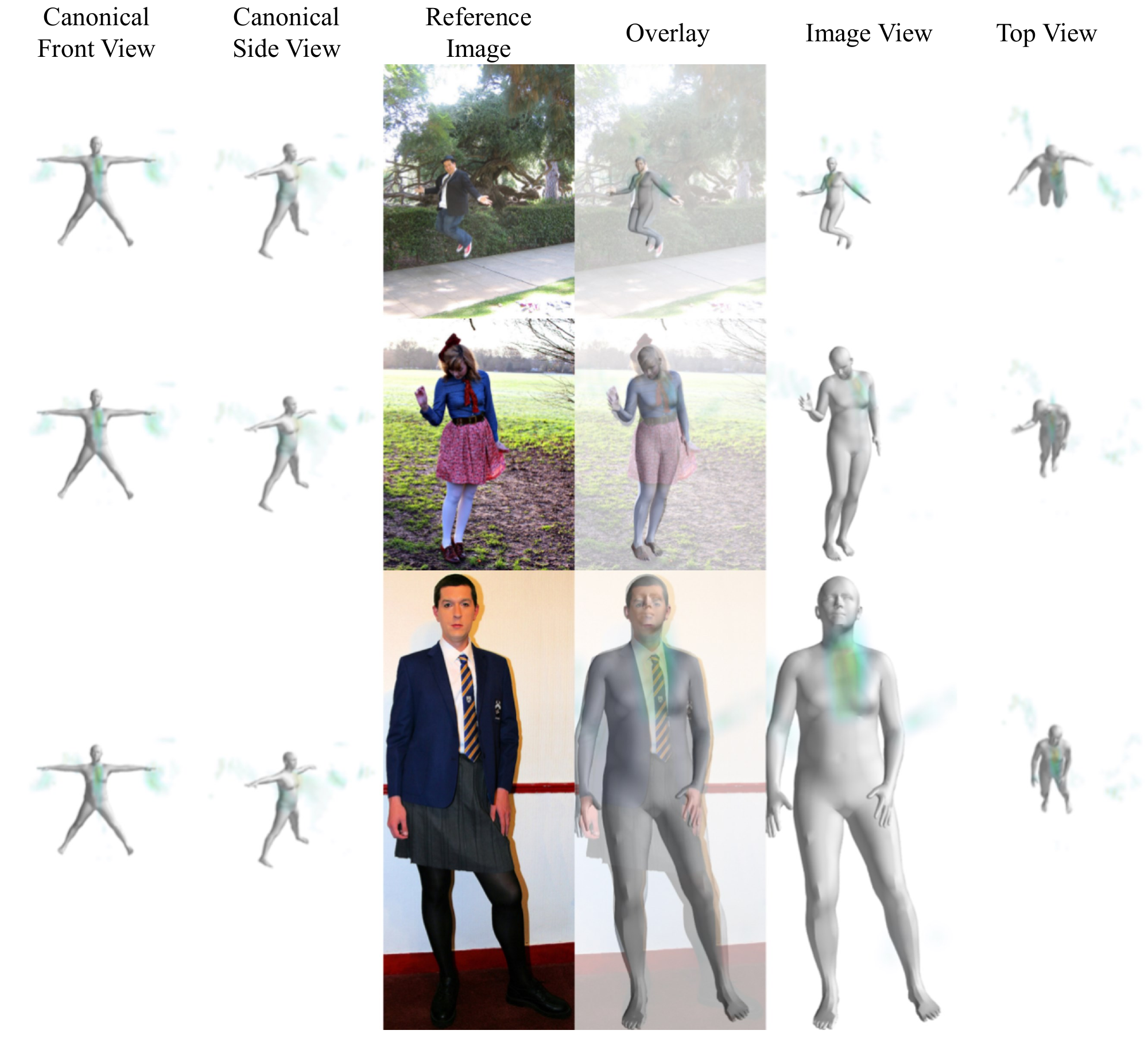}
    \caption{Qualitative results for category \textbf{tie}.}
    \label{fig:qual tie}
\end{figure*} \fi

\end{document}